\documentclass[10pt,twocolumn,letterpaper]{article}

\usepackage{wacv}              

\newcommand{\SC}[1]{\textcolor{blue}{\textbf{SC: #1}}}

\newcommand{\ourso}{DirectDrag (\textbf{ours})\xspace}
\newcommand{\ours}{DirectDrag\@\xspace}

\usepackage{xcolor}         
\usepackage[utf8]{inputenc} 
\usepackage[T1]{fontenc}    
\usepackage{url}            
\usepackage{booktabs}       
\usepackage{amsfonts}       
\usepackage{nicefrac}       
\usepackage{microtype}      

\usepackage{comment}

\RequirePackage{amsmath}
\RequirePackage{amssymb}
\RequirePackage{amsthm}
\RequirePackage{bm} 
\RequirePackage{url}
\usepackage{xspace}
\usepackage{multirow}
\usepackage{natbib}
\setcitestyle{square,comma,numbers,sort&compress}
\usepackage{graphicx}
\usepackage{caption,lipsum}

\usepackage{pifont}
\newcommand{\xmark}{\ding{55}}
\definecolor{checkmark}{HTML}{305AFF}
\definecolor{xmark}{HTML}{E62020}
\newcommand{\ccheck}{{\textcolor{checkmark}{\large\checkmark}}}
\newcommand{\ccross}{{\textcolor{xmark}{\large\xmark}}}

\usepackage{stfloats} 

\usepackage{pifont}
\usepackage{enumitem} 

\usepackage[usestackEOL]{stackengine}
\setstackgap{L}{\normalbaselineskip}

\usepackage{graphicx} 

\let\titleold\title
\renewcommand{\title}[1]{\titleold{#1}\newcommand{\thetitle}{#1}}
\def\maketitlesupplementary
   {
   \newpage
       \twocolumn[
        \centering
        \Large
        \textbf{\thetitle}\\
        \vspace{0.5em}Supplementary Material \\
        \vspace{1.0em}
       ] 
   }

\definecolor{wacvblue}{rgb}{0.21,0.49,0.74}
\usepackage[pagebackref,breaklinks,colorlinks,allcolors=wacvblue]{hyperref}

\usepackage[capitalize]{cleveref}
\crefname{section}{Sec.}{Secs.}
\Crefname{section}{Section}{Sections}
\Crefname{table}{Table}{Tables}
\crefname{table}{Tab.}{Tabs.}

\def\wacvPaperID{257} 
\def\confName{WACV}
\def\confYear{2025}

\begin{document}

\title{DirectDrag: High-Fidelity, Mask-Free, Prompt-Free Drag-based Image Editing via Readout-Guided Feature Alignment}

\author{
Sheng-Hao Liao$^1$\quad Shang-Fu Chen$^2$\quad Tai-Ming Huang$^2$\quad Wen-Huang Cheng$^2$\quad Kai-Lung Hua$^{1,3}$\\
$^1$National Taiwan University of Science and Technology\qquad $^2$National Taiwan University\\ $^3$Microsoft Taiwan
}

\twocolumn[{%
\renewcommand\twocolumn[1][]{#1}
\maketitle
\vspace{-30pt}
\begin{center}
    \centering
    \captionsetup{type=figure}
    \includegraphics[width=0.95\textwidth]{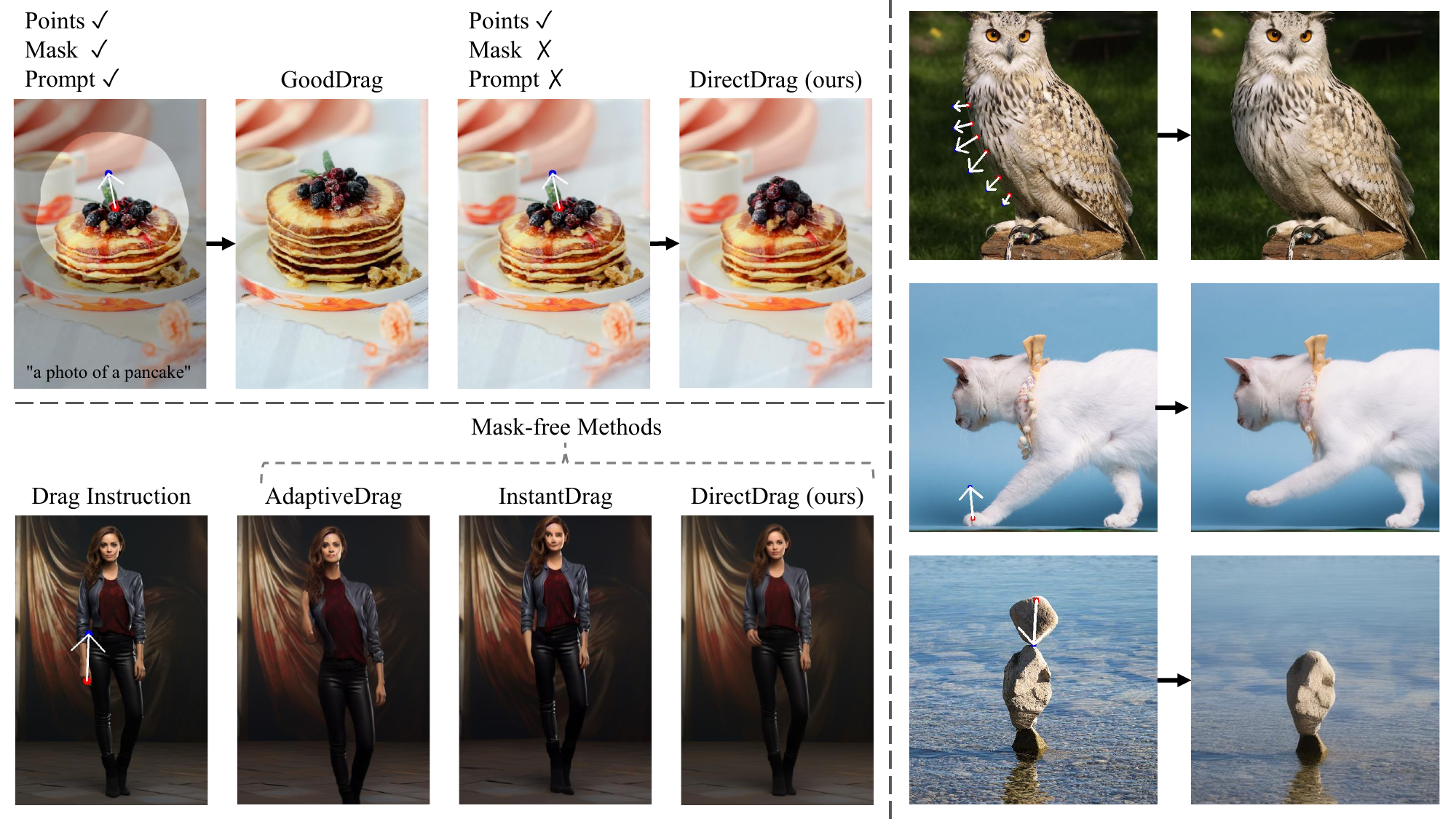}
    \vspace{-1em}
    \captionof{figure}{
    Up-Left: Existing methods such as GoodDrag~\cite{GoodDrag} require mask and prompt to assist the editing. Our \ours removes the dependency on mask and prompt, enabling more flexible editing while maintaining precise control. Bottom-left: Comparison with other manual mask-free methods, our method achieves more faithful and robust editing effects. Right: Additional qualitative results by \ours. Project Page: \url{https://frakw.github.io/DirectDrag/}.
    }
    \label{fig:teaser}
\end{center}
}]

\begin{abstract}
\label{sec:abstract}
Drag-based image editing using generative models provides intuitive control over image structures. However, existing methods rely heavily on manually provided masks and textual prompts to preserve semantic fidelity and motion precision. Removing these constraints creates a fundamental trade-off: visual artifacts without masks and poor spatial control without prompts. To address these limitations, we propose \ours, a novel mask- and prompt-free editing framework. \ours enables precise and efficient manipulation with minimal user input while maintaining high image fidelity and accurate point alignment. \ours introduces two key innovations. First, we design an Auto Soft Mask Generation module that intelligently infers editable regions from point displacement, automatically localizing deformation along movement paths while preserving contextual integrity through the generative model's inherent capacity. Second, we develop a Readout-Guided Feature Alignment mechanism that leverages intermediate diffusion activations to maintain structural consistency during point-based edits, substantially improving visual fidelity. Despite operating without manual mask or prompt, \ours achieves superior image quality compared to existing methods while maintaining competitive drag accuracy. Extensive experiments on DragBench and real-world scenarios demonstrate the effectiveness and practicality of \ours for high-quality, interactive image manipulation. Code is available at: \url{https://github.com/frakw/DirectDrag}.
\end{abstract}

\begin{figure*}[t!]
    \centering
    \includegraphics[width=\linewidth]{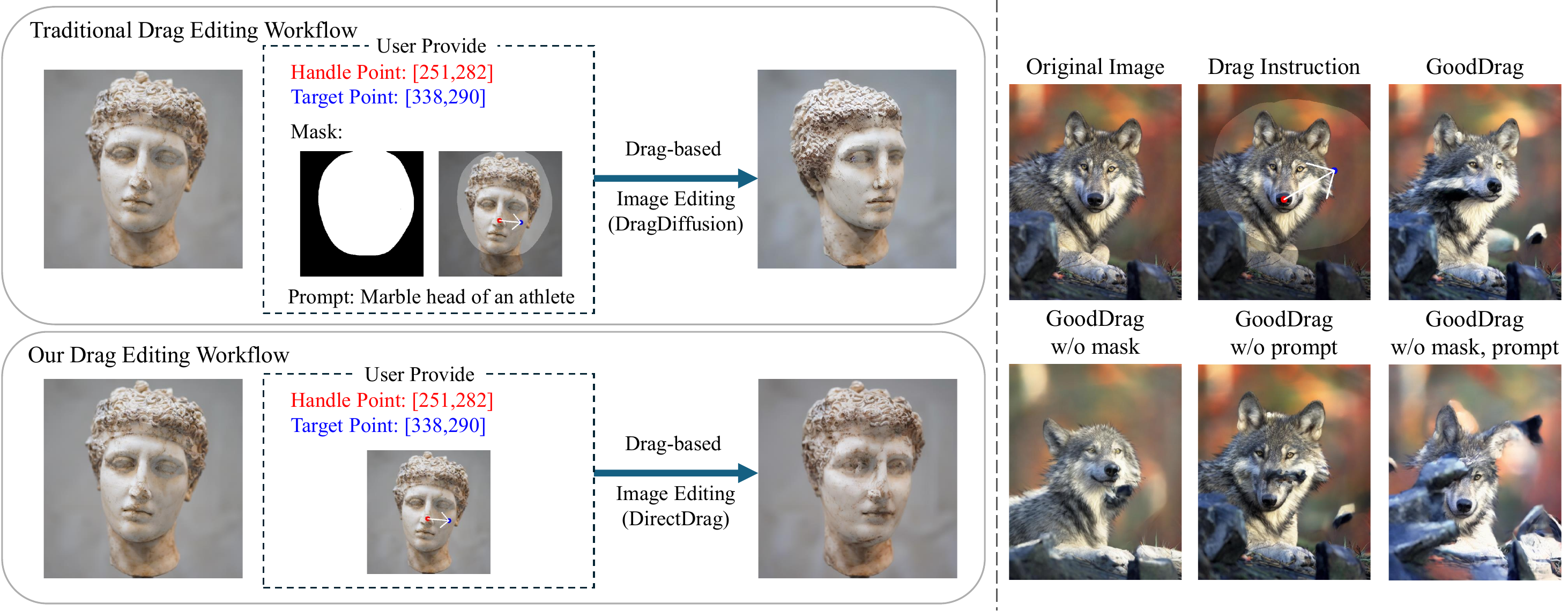}
    \caption{
    \textbf{Workflow Comparison.} \textbf{Left:} Traditional methods (e.g., DragDiffusion~\cite{DragDiffusion}, GoodDrag~\cite{GoodDrag}) rely on masks and prompts, increasing user burden. Our method simplifies the process by requiring only point inputs.  \textbf{Right:} Removing masks leads to distortion, while omitting prompts reduces accuracy. We demonstrate these effects on GoodDrag~\cite{GoodDrag} and also show the case without both inputs.
    }
    \label{fig:workflow_comparison}
\end{figure*}

\section{Introduction}\label{sec:introduction}
Drag-based image editing has become a powerful and intuitive way to manipulate visual content. With recent advances in diffusion-based generative models~\cite{DDPM, LDM}, this type of interaction has become increasingly precise and accessible. Unlike traditional text-to-image (T2I) methods~\cite{DALL-E-2, Imagen, GLIDE}, which rely on language to describe visual intentions, drag-based approaches provide direct and fine-grained control by allowing users to move a point from a source location to a desired target~\cite{DragGAN, DragDiffusion, DragonDiffusion}. This enables a wide range of image modifications, including facial expression editing, object repositioning, content resizing, restoration, and data augmentation. Many existing methods still require users to provide additional information, such as an editable region mask and a text prompt, to ensure accurate and semantically coherent results~\cite{FreeDrag, SDEDrag, GDrag, DADG}. These extra inputs, while helpful in guiding the editing process, create two major sources of annotation overhead and instability. First, manually drawing an appropriate mask becomes particularly difficult when users want to edit multiple parts of an image at once. In such cases, designing a precise mask is not only time-consuming but also prone to errors. Poorly drawn masks often result in unexpected distortions or artifacts. Second, cues are often difficult to formulate accurately, especially when images contain multiple semantically rich regions. Describing a complex visual environment in one sentence is extremely challenging, and even slight errors in the cues may mislead the diffusion model and lead to poor results. In some scenarios—such as medical imaging or technical illustrations—there may not even be suitable natural language to express the intended change, making prompt-based control impossible.
We find that removing the mask leads to noticeable loss of image fidelity (IF), while omitting the prompt significantly reduces point movement accuracy, reflected by increased mean distance (MD) scores. Therefore, eliminating these inputs, while desirable for simplifying user interaction, introduces real technical challenges. We illustrate these effects in Figure~\ref{fig:workflow_comparison}, where removing either the mask or the prompt leads to degraded visual quality or inaccurate drag results on a representative baseline (GoodDrag~\cite{GoodDrag}).
To address these issues, we present \textbf{\ours}, a novel drag-based editing framework that operates in a manual mask-free and prompt-free setting. Our method maintains high visual quality and competitive spatial precision, all while requiring only minimal and intuitive input: handle and target points. 

To achieve this, \ours integrates three core technical components:
\begin{itemize}
    \item An \textbf{Auto Soft Mask Generation} module that automatically infers editable regions based on point displacement. Rather than asking users to paint a mask manually, we localize deformation only along the path of movement, enhancing control where it matters most while relying on the generative model's capacity to preserve context elsewhere.
    \item A lightweight \textbf{Readout-Guided Feature Alignment} module that extracts intermediate diffusion features and aligns them based on spatial correspondence. This mechanism replaces the semantic guidance usually provided by prompt, helping the model maintain visual consistency and structure during editing.
    \item A \textbf{Latent Warpage Function}, adapted from prior work, which improves convergence and drag precision by initializing latent codes with a geometry-aware deformation. This component offers a prompt-free alternative to guide the optimization process toward semantically plausible outcomes.
\end{itemize}
Together, these components allow \ours to simplify the editing pipeline significantly. By removing the need for mask and prompt, we reduce the annotation burden and the risk of unstable or incorrect edits. As illustrated in Figure~\ref{fig:teaser}, our method outperforms existing manual mask-free approaches by producing more faithful and robust edits, even with minimal inputs. Despite having fewer user-provided signals, our approach achieves higher image fidelity than strong baseline. Although there is a slight trade-off in drag accuracy compared to full-input systems, the difference remains small. This suggests that our framework provides a favorable balance between usability and performance. We validate the effectiveness of \ours through extensive experiments on DragBench and real-world images, confirming its potential for practical and scalable interactive editing.

\section{Related Work}\label{sec:related_work}

\subsection{Generative Image Models and Image Editing}
Generative image models, particularly GANs and diffusion models, have significantly enhanced image synthesis and editing capabilities. GANs~\cite{GAN, StyleGAN2} provide fast generation, but stable reversible editing is often difficult to achieve. Diffusion models~\cite{DDPM, LDM, Imagen} show outstanding fidelity through iterative denoising of latent codes. These models form the basis of interactive image editing applications. Image editing techniques can be divided into content-aware and content-free methods:
Content-Aware Editing includes object manipulation, spatial transformation, inpainting, and style transfer. Text-prompted editing methods (e.g., InstructPix2Pix~\cite{InstructPix2Pix, Imagen}) and user-guided approaches fall into this category.
Content-Free Editing focuses on customization using user-specified images or attributes. Examples include subject-driven personalization (e.g., DreamBooth~\cite{DreamBooth}) and attribute-driven fine-tuning.

\subsection{Drag-based Image Editing}
Drag-based image editing methods enable users to control image structures by dragging specific points to target locations. DragGAN~\cite{DragGAN} first proposed a latent code optimization framework with point tracking based on GANs, but struggled with generalizing to real-world inputs. DragDiffusion~\cite{DragDiffusion} and DragonDiffusion~\cite{DragonDiffusion} extended this paradigm to diffusion models, improving structural manipulation and semantic controllability through prompt conditioning and denoising-based alignment.

Subsequent methods aimed at improving editing quality and robustness. DragNoise~\cite{DragNoise} reduces cost by optimizing U-Net bottleneck features. GoodDrag~\cite{GoodDrag} alternates dragging and denoising to prevent error accumulation. GDrag~\cite{GDrag} is training-free, addressing intention and content ambiguity via atomic manipulations and dense trajectories. FlowDrag~\cite{FlowDrag} improves geometric consistency with 3D mesh-guided flow fields. DragLoRA~\cite{DragLoRA} enhances precision and efficiency through online LoRA adaptation with adaptive motion supervision.

Other works focus on enhancing editing efficiency. DiffEditor~\cite{DiffEditor} reduces optimization time by decreasing the number of diffusion steps. FastDrag~\cite{FastDrag} uses a one-step feed-forward generation approach for instant edits. LightningDrag~\cite{LightningDrag} treats editing as conditional generation trained on large-scale video data for fast, accurate results. EEdit~\cite{EEdit} accelerates editing by reducing spatial and temporal redundancy through region caching and inversion step skipping.

\subsection{Manual Mask-Free Drag-Based Image Editing}
Recent works have proposed removing manually provided masks to simplify the drag editing pipeline while preserving semantic and structural control. EasyDrag\cite{EasyDrag} focuses on user-friendliness by eliminating the need for masks and tuning procedures such as LoRA\cite{LoRA}. It leverages pretrained diffusion models without architectural modifications and achieves better editing precision and visual quality than DragDiffusion~\cite{DragDiffusion}. However, it still requires a text prompt to maintain semantic guidance, which limits usability in prompt-free scenarios. In addition, EasyDrag relies on ControlNet~\cite{ControlNet}, which introduces considerable memory overhead during inference.

InstantDrag~\cite{InstantDrag} improves editing speed by introducing an optimization-free pipeline that takes only an image and a drag instruction as input. It uses a drag-conditioned optical flow network followed by a flow-guided diffusion model to achieve fast and realistic edits. While it avoids mask and prompt, InstantDrag must retrain a dedicated diffusion model on large-scale video data, significantly increasing parameter count and training cost. Moreover, it often requires multiple drag instructions to produce stable results, reducing its effectiveness in sparse user-interaction settings.
AdaptiveDrag~\cite{AdaptiveDrag} introduces automatic mask generation using superpixel segmentation by SAM2~\cite{SAM2} and incorporates semantic-aware latent optimization guided by adaptive steps and a specialized loss. Although it improves localization accuracy and generalization across categories, AdaptiveDrag depends on external segmentation models and still requires textual prompt for semantic alignment, resulting in additional computational overhead.

While these methods effectively reduce the need for manual mask input, they either rely on prompt, introduce heavy architectural modifications, or require extra modules such as segmentation or flow estimation. In contrast, \textbf{\ours} adopts a lightweight and manual mask-free and prompt-free framework that maintains high image fidelity and competitive drag precision. It achieves this through automatic soft mask generation, readout-guided feature alignment, and latent warpge function introducing only a minimal auxiliary module, far more efficient and compact than the large-scale components used in existing approaches.

\begin{figure*}[htbp]
    \centering
    \includegraphics[width=0.9\linewidth]{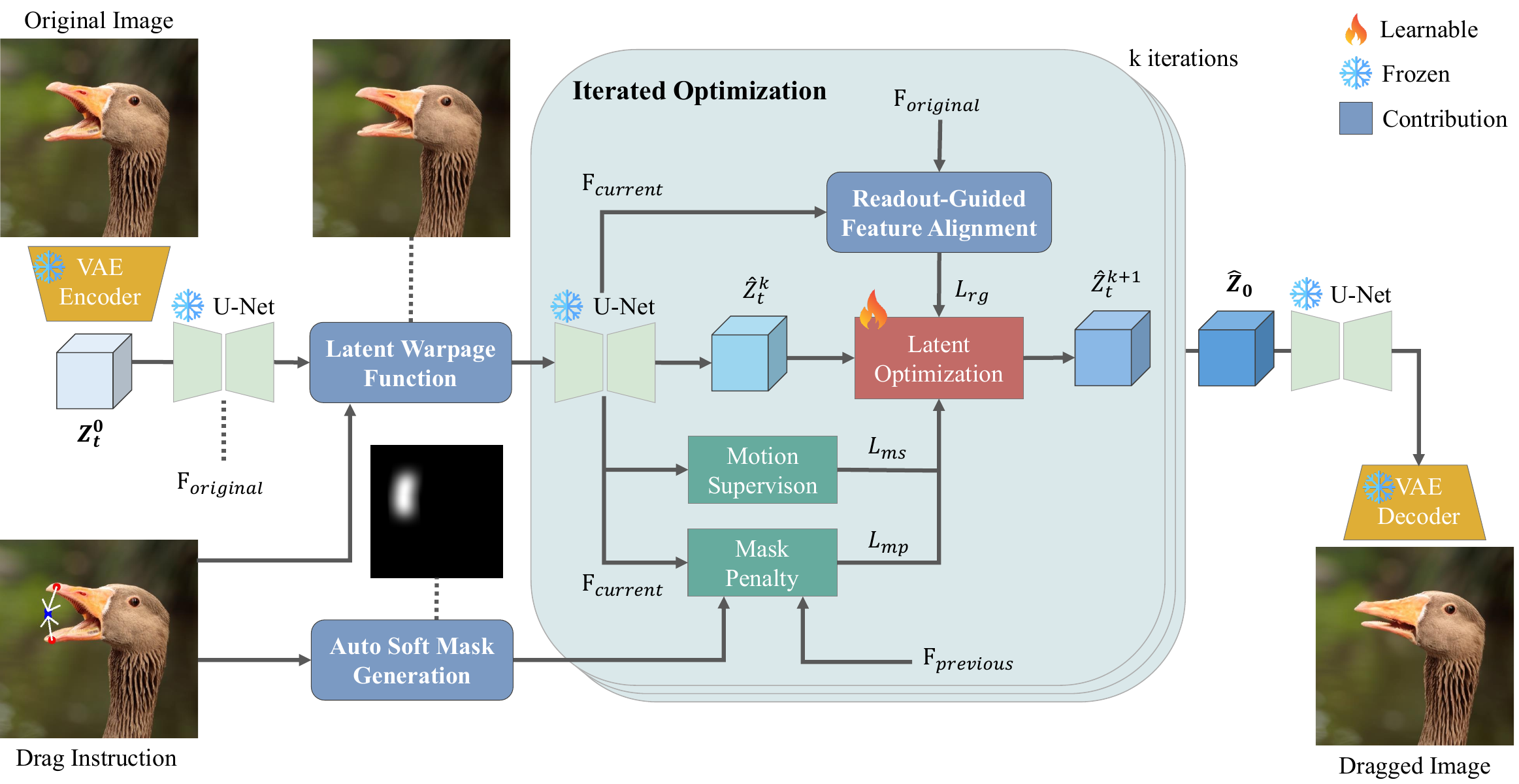}
    \caption{
    \textbf{Overview of the proposed \ours framework}. Given an input image and point pairs, we apply DDIM inversion to obtain latent codes, initialize editing via latent warpage function and generate soft mask, then iteratively apply drag and denoising guided by motion supervision and feature alignment.
    }
    \label{fig:framework_overview}
\end{figure*}

\section{Method}\label{sec:method}

\subsection{Overview}\label{subsec:overview}

We propose \textbf{\ours}, a manual mask-free and prompt-free framework for drag-based image editing. Unlike previous diffusion-based methods~\cite{DragDiffusion, DragonDiffusion, StableDrag, GoodDrag}, which rely on hand-crafted mask or prompt, our method simplifies the pipeline while preserving editing quality.
As shown in Figure~\ref{fig:framework_overview}, the process begins by applying DDIM inversion~\cite{DDIM} to encode the input image into latent space. A geometry-aware latent warpage function (LWF) initializes the latent code, and an auto soft mask generation module estimates the editable region based on point displacement—removing the need for manual masks.
We adopt the AlDD strategy~\cite{GoodDrag} (Alternating-Drag-and-Denoising) to optimize the latent representation iteratively. During each step, drag loss encourages point movement, while our readout-guided Feature alignment module extracts intermediate diffusion features to maintain visual consistency. These components work together to preserve fidelity and precision even without prompts or segmentation inputs.

Compared to prior work that introduces architectural changes~\cite{InstantDrag} or external segmentation tools~\cite{AdaptiveDrag}, \ours remains lightweight and modular, while achieving strong fidelity and alignment performance across diverse examples.

\subsection{Latent Diffusion and DDIM Inversion}

Denoising Diffusion Probabilistic Models (DDPMs)~\cite{DDPM} have demonstrated strong generative capabilities by modeling the image generation process as a gradual denoising of random noise. However, operating directly in pixel space is computationally expensive. To improve efficiency, Latent Diffusion Models (LDMs)~\cite{LDM} encode the image $\mathbf{x}_0$ into a lower-dimensional latent representation $\mathbf{z}_0 = \mathcal{E}(\mathbf{x}_0)$ using a pretrained VAE encoder $\mathcal{E}$. The diffusion process is then carried out in the latent space as a Markov chain over $T$ timesteps, where the marginal likelihood is expressed as:

\begin{equation}
    p_\theta(\mathbf{z}_0) = \int p_\theta(\mathbf{z}_{1:T}) \, d\mathbf{z}_{1:T},
\end{equation}
where each latent variable $\mathbf{z}_t$ is obtained by progressively adding Gaussian noise to $\mathbf{z}_0$ using a forward process defined as:

\begin{equation}
    \mathbf{z}_t = \sqrt{\bar{\alpha}_t} \, \mathbf{z}_0 + \sqrt{1 - \bar{\alpha}_t} \, \boldsymbol{\epsilon}, \quad \boldsymbol{\epsilon} \sim \mathcal{N}(0, \mathbf{I}),
\end{equation}
where $\bar{\alpha}_t$ denotes the cumulative product of noise schedule coefficients up to timestep $t$.

To enable editing from real images, we adopt deterministic DDIM inversion~\cite{DDIM}, which reverses the diffusion process to recover latent trajectories. This allows us to initialize the editing process from a clean latent code $\mathbf{z}_0$ without requiring random sampling. Since our method does not rely on prompts, DDIM inversion is performed in a prompt-free setting, enabling faithful reconstructions and providing a robust starting point for subsequent drag-based manipulation.

\begin{figure*}[htbp]
    \centering
    \includegraphics[width=\linewidth]{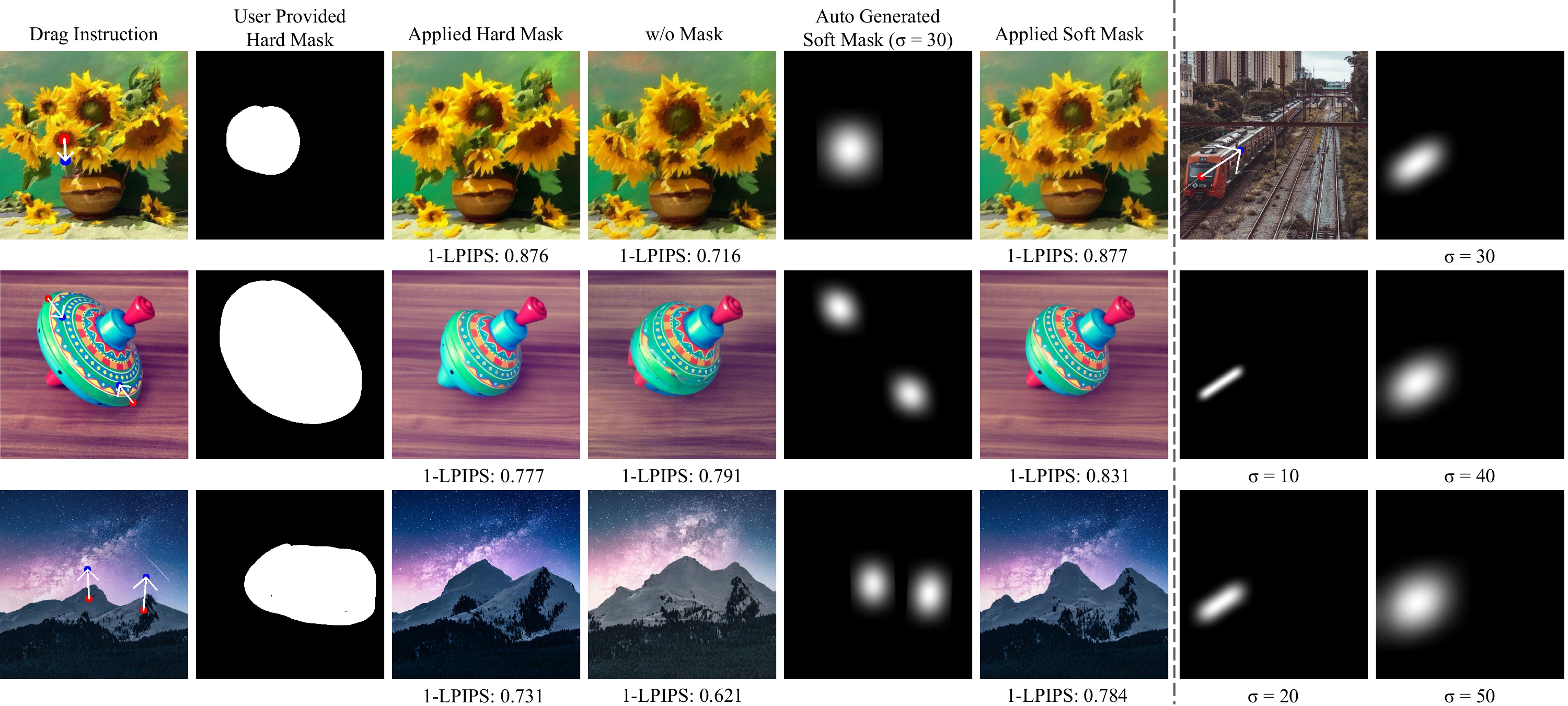}
    \caption{\textbf{Effect of our Soft Mask.} Left: Compared to no masking and user provide hard mask, applying the generated soft mask significantly improves visual fidelity and structure preservation, as reflected by higher image fidelity scores (1-LPIPS$\uparrow$). Right: Visualization of soft masks under different drag configurations and Gaussian widths~($\sigma$), illustrating their adaptiveness to motion magnitude and direction.}
    \label{fig:soft_mask}
    \vspace{-6pt}
\end{figure*}

\subsection{Drag-based Image Editing}\label{subsec:prelim_drag}

Our method builds upon prior drag-based diffusion editing approaches~\cite{DragDiffusion, GoodDrag}, where user-specified handle points are iteratively moved toward target locations by optimizing latent features in the diffusion model. To guide this deformation process, we incorporate three key components: motion supervision, alternating drag and denoising, and feature-based point tracking.


\noindent\textbf{Motion Supervision.} We adopt a multi-step motion supervision loss to encourage the features at displaced handle points to match those at their original locations. This supervision helps align internal features with the intended motion trajectory:

\begin{equation}
\mathcal{L}_{\text{ms}} = 
\sum_{i=1}^{n} \sum_{q} \left\| \mathcal{F}_{q + d_i}(\hat{\mathbf{z}}_t^k, \hat{\mathbf{c}}^k) 
- \text{sg}(\mathcal{F}_{q}(\hat{\mathbf{z}}_t^k, \hat{\mathbf{c}}^k)) \right\|_1,
\end{equation}
where $\mathcal{F}_{q}$ denotes the U-Net features extracted at location $q$, and $d_i$ is the displacement vector of the $i$-th handle point.

\noindent\textbf{AlDD Optimization Schedule.}
To prevent noise accumulation and preserve global image structure, we adopt the AlDD schedule proposed in GoodDrag~\cite{GoodDrag}. Rather than performing continuous updates in the latent space, AlDD interleaves $B$ drag steps with periodic denoising steps. This scheduling helps retain proximity to the image manifold and stabilizes optimization. At each drag step, we apply a patch-level alignment loss:

\begin{equation}
\mathcal{L}_{\text{drag}} = 
\sum_i \left\| \mathcal{F}_{\Omega(\mathbf{p}_i + \delta \mathbf{p}_i)} - \text{sg}(\mathcal{F}_{\Omega(\mathbf{p}_i)}) \right\|_1,
\end{equation}
where $\Omega(\cdot, r_1)$ extracts a spatial patch of radius $r_1$, and $\delta \mathbf{p}_i^k$ is the displacement from the initial handle position to its target.

\noindent\textbf{Point Tracking.}
We also incorporate the point tracking mechanism from GoodDrag~\cite{GoodDrag} to maintain semantic consistency throughout the editing trajectory. Instead of keeping handle points fixed across iterations, we dynamically update each point's position by matching its initial diffusion features with features from nearby locations in the current timestep. This allows the model to follow the semantic content even as the image structure evolves during optimization. The detailed formulation of this tracking algorithm is provided in the supplementary material.

Together, motion supervision, AlDD scheduling, and feature-based tracking form the core optimization loop that enables precise point-based editing while preserving image quality and structural coherence.

\begin{figure*}[htbp]
    \centering
    \includegraphics[width=0.95\textwidth]{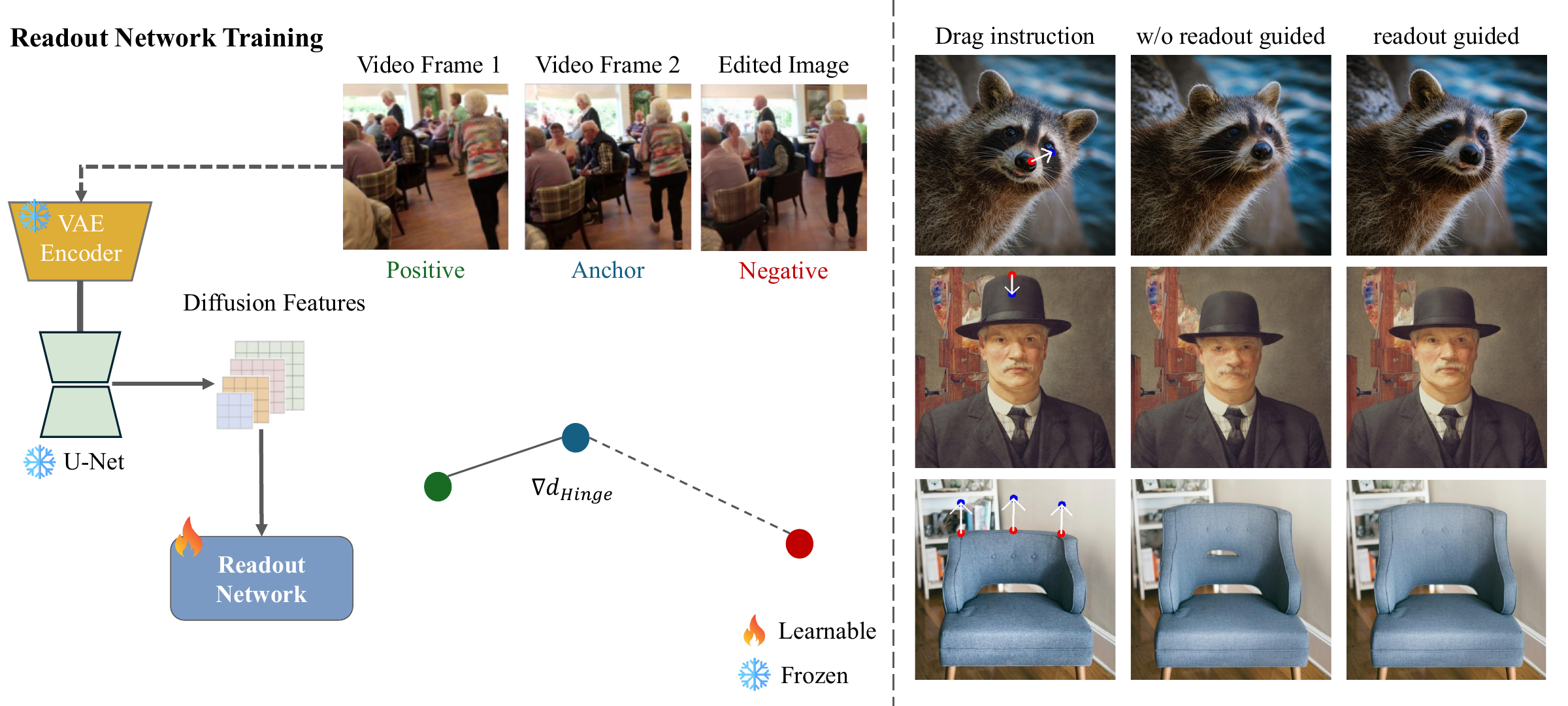}
    \caption{
    \textbf{Readout Network Training and Effect.} Left: We train the readout network using a triplet loss on diffusion features extracted from video frames (anchor, positive) and edited images (negative). Right: Incorporating readout guidance preserves appearance details and improves structural consistency during dragging.
    }
    \label{fig:readout}
    \vspace{-6pt}
\end{figure*}

\subsection{Auto Soft Mask Generation}

In drag-based editing, prior methods often rely on user-provided hard mask to confine deformation. However, even with these mask, diffusion models tend to produce unintended changes in unrelated regions due to weak spatial constraints. In practice, omitting mask altogether leads to even more severe artifacts, such as missing objects, hallucinated structures, or drastic changes in color and composition—as shown in Fig.~\ref{fig:soft_mask}.

To improve usability while reducing over-editing, we propose to generate a soft spatial mask $M \in [0,1]^{H \times W}$ directly from the drag instructions. This removes the burden of manual annotations and ensures localized structural control. Specifically, for each handle–target pair $(\mathbf{h}_i, \mathbf{t}_i)$ with coordinates $(x_0, y_0)$ and $(x_1, y_1)$, we interpolate $N = \max(|x_1 - x_0|, |y_1 - y_0|) + 1$ points along the linear path connecting them:

\begin{align}
\tilde{M}(x_k, y_k) &= 1, \quad \text{where} \notag \\
(x_k, y_k) &= \left\lfloor (1 - \alpha_k)(x_0, y_0) + \alpha_k(x_1, y_1) \right\rceil, \\
\alpha_k &= \frac{k}{N-1} = \frac{k}{\max(|x_1 - x_0|, |y_1 - y_0|)}.
\end{align}

We accumulate $\tilde{M}$ from all point pairs, then apply a Gaussian filter followed by normalization to form the final soft mask $M$:

\begin{equation}
M = \frac{\text{GaussianBlur}(\tilde{M}, \sigma)}{\max\left(\text{GaussianBlur}(\tilde{M}, \sigma)\right)}.
\end{equation}

The resulting soft mask softly highlights the regions along dragging trajectories, enforcing smooth, localized constraints without introducing sharp editing boundaries. While this design significantly reduces unintended edits, it has its limitations: the linear interpolation path may not fully cover the deformable object, especially for complex geometries. Nevertheless, we argue that the primary role of a mask is to localize major structural changes—not to precisely capture every affected pixel. In fact, over-constraining the optimization via strict loss masking can conflict with the global nature of latent updates in diffusion models, sometimes degrading drag precision instead of improving it. Our lightweight mask acts as a guiding prior, with finer control delegated to subsequent alignment mechanisms.

\subsection{Readout-Guided Feature Alignment}

Although the soft mask improves visual fidelity and local stability, it often fails to suppress subtle background artifacts or hallucinated textures, as illustrated in Fig.~\ref{fig:readout}. To address this, we incorporate a feature alignment mechanism based on Diffusion Hyperfeatures~\cite{DiffusionHyperfeatures} and Readout Guidance~\cite{ReadoutGuidance}.

\noindent\textbf{Readout Network.}
Following Luo \etal~\cite{ReadoutGuidance}, we use a lightweight readout network trained to extract appearance-preserving features from intermediate U-Net layers of a frozen denoiser. Supervision is provided via a triplet loss:

\begin{equation}
\begin{split}
\mathcal{L}_{\text{triplet}} = \max\big(0,\ 
& D(F(I_a), F(I_p)) \\
& - D(F(I_a), F(I_n)) + \delta \big)
\end{split}
\end{equation}
where $F(\cdot)$ is the readout head output, $D$ is cosine distance, and $I_p$, $I_n$ are positive and negative samples. Negative examples are generated by SDEdit~\cite{SDEdit}, which perturbs appearance while preserving structure. Readout Guidance~\cite{ReadoutGuidance} use training data from the DAVIS dataset~\cite{DAVIS}. 

\noindent\textbf{Inference-Time Guidance.}
During editing, we extract intermediate features from the original image $\mathbf{z}_t^\text{0}$ (before any dragging) and use them as the reference for appearance alignment. For each optimization step, the current latent $\bar{\mathbf{z}}_t$ is passed through the readout network, and the following loss is applied:

\begin{equation}
\mathcal{L}_{\text{rg}} = \| F(\bar{\mathbf{z}}_t^\text{k}) - F(\mathbf{z}_t^\text{0}) \|_2^2,
\end{equation}

\noindent where $F(\cdot)$ denotes the readout network's output from selected U-Net layers (e.g., down3 to up2). This encourages the edited latent to stay visually close to the original appearance, mitigating hallucination and identity drift.
Unlike Readout Guidance~\cite{ReadoutGuidance}, which is designed for one-shot diffusion and prone to hallucinations, our approach integrates readout features into a multi-step optimization framework. This allows better convergence and reduces artifacts, especially in challenging scenes. The guidance is effective without modifying the diffusion backbone, introducing only minor overhead while improving appearance stability.

\begin{table*}[htbp]
\vspace{-2em}
\centering
\setlength{\tabcolsep}{5pt}
\begin{tabular}{l|c|cc|ccc|cc}
\toprule
\textbf{Method} & \textbf{Venue} & \textbf{Mask} & \textbf{Prompt} & \textbf{IF$\uparrow$} & \textbf{CLIP SIM$\uparrow$} & \textbf{MD$\downarrow$} & \Centerstack{\textbf{Model} \\ \textbf{Params}} & \Centerstack{\textbf{Tuning} \\ \textbf{Params}} \\
\midrule
DragDiffusion~\cite{DragDiffusion}     & CVPR'24    & \ding{51} & \ding{51} & 0.883 & 0.977 & 32.87 & 865M & 0.07M \\
FreeDrag~\cite{FreeDrag}     & CVPR'24    & \ding{51} & \ding{51} & 0.897 & 0.977 & 33.82 & 865M & 0.07M \\
DiffEditor~\cite{DiffEditor}     & CVPR'24    & \ding{51} & \ding{51} & 0.877 & 0.966 & 31.70 & 865M & 0.07M \\
DragNoise~\cite{DragNoise}         & CVPR'24    & \ding{51} & \ding{51} & 0.899 & 0.972 & 37.92 & 865M & 0.33M \\
FastDrag~\cite{FastDrag}          & NeurIPS'24 & \ding{51} & \ding{51} & 0.859 & 0.963 & 32.66 & 865M & 0 \\
GoodDrag~\cite{GoodDrag}          & ICLR'25    & \ding{51} & \ding{51} & 0.869 & 0.977 & 25.28 & 865M & 0.07M \\
DragText~\cite{DragText}          & WACV'25    & \ding{51} & \ding{51} & 0.870 & 0.971 & 34.25 & 865M & 0.12M \\
LightningDrag~\cite{LightningDrag}     & ICML'25    & \ding{51} & \ding{51} & 0.881 & 0.970 & 29.95 & 933M & 933M \\
\cmidrule(lr){1-9}
\multicolumn{9}{c}{\textit{Manual Mask-free methods}} \\
\cmidrule(lr){1-9}
EasyDrag*~\cite{EasyDrag}          & CVPR'24    & \ding{55} & \ding{51} & 0.882 & -- & 34.44 & 1770M & \textbf{0.07M} \\
Readout Guidance~\cite{ReadoutGuidance}          & CVPR'24    & \ding{55} & \ding{55} & 0.867 & 0.951 & 55.12 & \textbf{871M} & \underline{5.97M} \\
AdaptiveDrag~\cite{AdaptiveDrag}      & ArXiv'24   & \ding{55} & \ding{51} & 0.867 & 0.975 & 33.94 & 1168M & \textbf{0.07M} \\
InstantDrag~\cite{InstantDrag}       & {\fontsize{7pt}{8pt}\selectfont SIGGRAPH Asia'24}   & \ding{55} & \ding{55} & 0.878 & 0.968 & \underline{30.41} & 914M & 914M \\
$\textbf{\ourso}_\text{w/o LWF}$   & --         & \ding{55} & \ding{51} & \textbf{0.918} & \textbf{0.982} & 31.91 & \textbf{871M} & \underline{5.97M} \\
\textbf{\ourso}   & --         & \ding{55} & \ding{55} & \underline{0.891} & \underline{0.976} & \textbf{29.65} & \textbf{871M} & \underline{5.97M} \\
\bottomrule
\end{tabular}
\vspace{-.5em}
\caption{
\textbf{Quantitative evaluation} on the DragBench~\cite{DragDiffusion} dataset. IF = 1 - LPIPS. CLIP SIM = CLIP~\cite{CLIP} Similarity. MD = Mean Distance.  
\ding{51}: Required, \ding{55}: Not Required. LWF: Latent Warpage Function.  Model Params: Total parameters used in model. Tunning Params: Parameters require to training in correspond method. * means scores are taken from the another publication.
}
\vspace{-0.5em}
\label{tab:dragbench}
\end{table*}

\begin{table}[t]
\centering
\small
\resizebox{\linewidth}{!}{ 
\begin{tabular}{lcccc|c}
\toprule
Method & $\gamma=1$ & $\gamma=5$ & $\gamma=10$ & $\gamma=20$ & GScore $\uparrow$\\
\midrule
DragDiffusion~\cite{DragDiffusion}       & 0.1189 & 0.1101 & 0.0979 & 0.0924 & 6.90 \\
SDE-Drag~\cite{SDEDrag}  & 0.1571 & 0.1437 & 0.1291 & 0.1143 & 5.38 \\
GoodDrag~\cite{GoodDrag}   & \textbf{0.0696} & \textbf{0.0673} & \textbf{0.0642} & \textbf{0.0623} & \textbf{7.94} \\
\textbf{\ourso}     & \underline{0.1124} & \underline{0.1044} & \underline{0.0978} & \underline{0.0916} & \underline{6.95}  \\
\bottomrule
\end{tabular}
}
\vspace{-.5em}
\caption{Quantitative evaluation of drag accuracy in terms of \textbf{DAI} and \textbf{GScore} on Drag100.  
Lower values indicate more accurate drag editing. Other scores are taken from GoodDrag~\cite{GoodDrag}.}
\vspace{-1.2em}
\label{tab:dai_gscore}
\end{table}

\subsection{Latent Warpage Function}
To initialize the latent with geometry-aware deformation, we adopt the latent warpage function (LWF) from FastDrag~\cite{FastDrag}. For each masked pixel $p_j$ in latent space, its displacement $\mathbf{v}_j$ is computed as a weighted combination of drag vectors $\mathbf{d}_i = e_i - s_i$:

\begin{equation}
\mathbf{v}_j = \sum_{i=1}^{k} w_j^i \cdot \lambda_j^i \cdot \mathbf{d}_i,
\end{equation}
where $w_j^i$ is the inverse distance weight to handle $s_i$, and $\lambda_j^i$ is a stretch factor based on geometric intersections.  

Unlike the original latent warpage function, which often over-applies displacement and harms fidelity, we scale the drag vector with a ratio $\rho$:

\begin{equation}
\mathbf{d}_i' = \rho \cdot (e_i - s_i),
\end{equation}
producing a gentler shift in latent space. This mitigates early semantic drift and improves convergence. Empirically, this initialization reduces mean distance error and enables more stable drag optimization in subsequent steps.

\section{Experiments}\label{sec:exp}

\subsection{Implementation Details}

We build on Stable Diffusion v1.5~\cite{LDM} and run all experiments on single NVIDIA RTX 4090. Our pipeline follows DDIM inversion with 50 inference steps and guidance scale 1.0. We highlight three key settings: (1) Soft Mask: Gaussian blur with $\sigma = 30$. (2) Readout-Guided Weight: The readout guidance loss is scaled by 350 before adding to the main objective. (3) Latent Warpage Function: To reduce over-drag during initialization, we apply 15\% of the displacement vector from handle to target. All other parameters follow settings from baseline (GoodDrag~\cite{GoodDrag}). For an input image of $512 \times 512$ with a single drag instruction, it takes approximately 20 seconds to train LoRA and around 50 seconds for editing inference, utilizing about 13 GB of VRAM.

\subsection{Quantitative Evaluation}

We evaluate on DragBench~\cite{DragDiffusion} using (1) 1-LPIPS~\cite{LPIPS} for perceptual similarity, (2) CLIP~\cite{CLIP} Similarity for semantic consistency, and (3) MD~\cite{DragGAN} for dragging accuracy using DIFT~\cite{DIFT}.
As shown in Table~\ref{tab:dragbench}, \ours perfrom \textbf{state-of-the-art} result in manual mask-free methods. Despite working in minimal input conditions, \ours matches or exceeds mask-based and prompt-based methods in image fidelity and drag accuracy. We also test our method on Drag100 dataset by DAI and GScore metrics, see Table~\ref{tab:dai_gscore}.

\begin{figure*}[htbp]
    \vspace{-2em}
    \centering
    \includegraphics[width=0.8\linewidth]{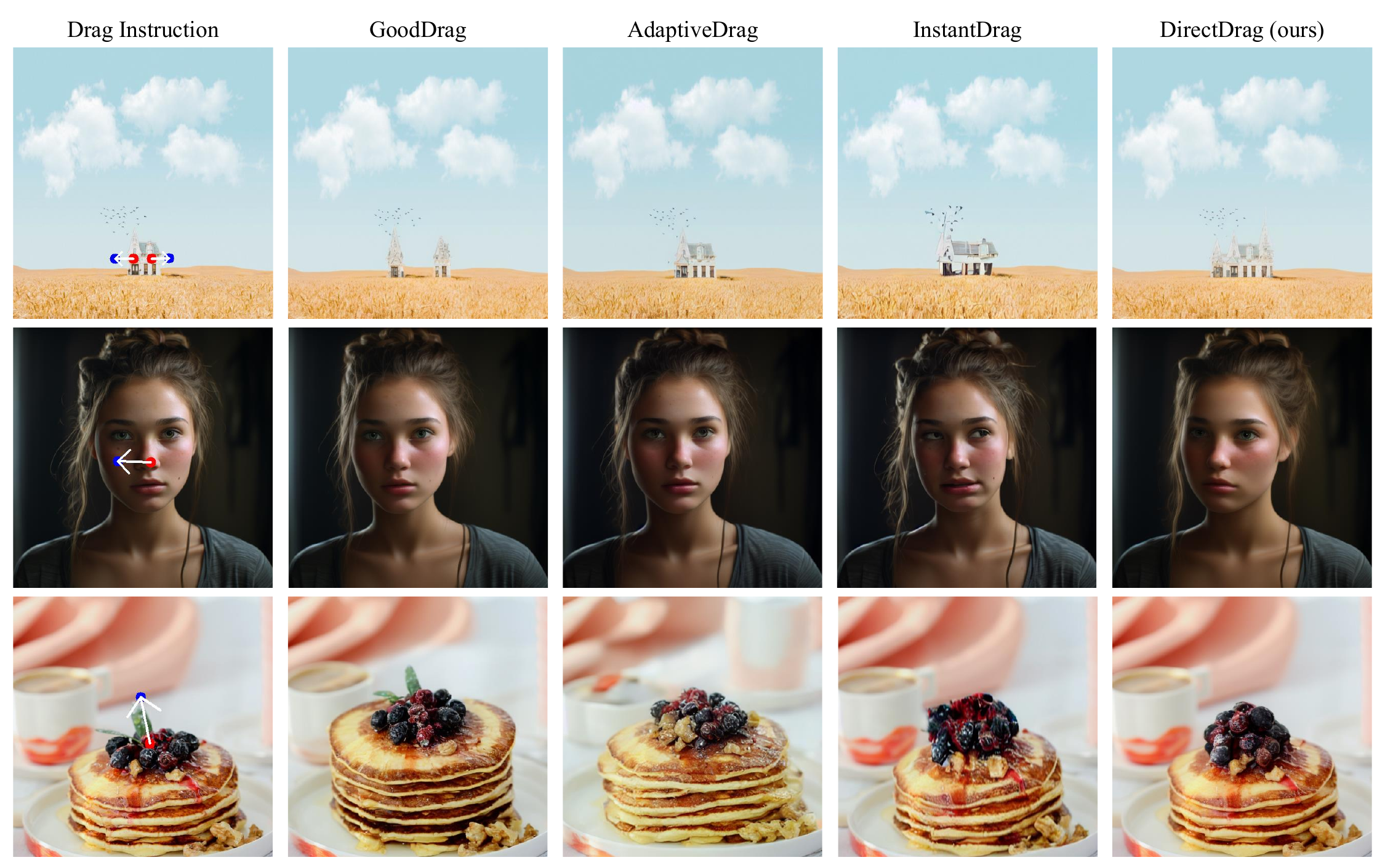}
    \caption{
    \textbf{Qualitative comparison}. Compared to the baseline (\textit{GoodDrag}~\cite{GoodDrag}) and manual mask-free methods (\textit{AdaptiveDrag}~\cite{AdaptiveDrag}, \textit{InstantDrag}~\cite{InstantDrag}), our method \textit{\ours}
    }
    \label{fig:qualitative}
    \vspace{-1em}
\end{figure*}

\subsection{Qualitative Results}

Fig.~\ref{fig:qualitative} compares \ours to GoodDrag~\cite{GoodDrag} (baseline with mask and prompt) and two manual mask-free methods, AdaptiveDrag~\cite{AdaptiveDrag} and InstantDrag~\cite{InstantDrag}. While the latter often suffers from distortions or incomplete motion, our method achieves more accurate and stable edits.
Across diverse cases—motion, face, and object deformation—\textit{\ours} maintains background consistency and visual detail, confirming its advantage in prompt-free and manual mask-free editing.

\begin{table}[t]
\centering
\small
\resizebox{\linewidth}{!}{ 
\begin{tabular}{lccc|ccc}
\toprule
Method & SM & RG & LWF & IF$\uparrow$ & \Centerstack{CLIP \\ SIM}$\uparrow$ & MD$\downarrow$ \\
\midrule
Baseline &       &       &       & 0.789 & 0.963 & 24.74 \\
\midrule
+ Soft Mask       & \ding{51} &       &       & 0.895 & 0.979 & 31.35 \\
+ Readout Guided   & \ding{51} & \ding{51} &       & \textbf{0.918} & \textbf{0.982} & 33.75 \\
+ Readout Guided $_\text{+prompt}$   & \ding{51} & \ding{51} &       & \textbf{0.918} & \textbf{0.982} & 31.91 \\
+ Latent Warpage     & \ding{51} & \ding{51} & \ding{51} & 0.891 & 0.976 & 29.65  \\
+ Latent Warpage $_\text{+prompt}$     & \ding{51} & \ding{51} & \ding{51} & 0.891 & 0.975 & \textbf{29.18}  \\
\bottomrule
\end{tabular}
}
\vspace{-.5em}
\caption{\textbf{Ablation study of \textit{\ours}}. Baseline indicates GoodDrag~\cite{GoodDrag} without mask and prompt.}
\vspace{-1em}
\label{tab:ablation}
\end{table}

\subsection{Ablation Study}

Table~\ref{tab:ablation} shows the impact of each component in \textit{\ours}. The soft mask significantly improves visual fidelity, while readout guidance helps preserve appearance but slightly reduces motion accuracy. Latent warpage function improves spatial precision with minimal degradation in image quality. We also tested a variant using prompt conditioning, showing that our latent warpage function can effectively replace prompt for improving drag accuracy. Overall, our final setup offers the best trade-off between fidelity and accuracy in a manual mask-free and prompt-free setting.

\section{Conclusion}

We presented \textit{\ours}, a lightweight framework for drag-based image editing that operates without manual mask or prompt. By integrating automatic soft mask generation, readout-guided feature alignment, and a latent warpage function, our method achieves high visual fidelity and competitive dragging accuracy. Extensive experiments demonstrate that \textit{\ours} provides a practical and effective solution for intuitive image manipulation, balancing usability, precision, and quality.

\clearpage
{\small
\bibliographystyle{ieee_fullname}
\bibliography{egbib}

@String(CVPR= {IEEE Conf. Comput. Vis. Pattern Recog.})

@String(ICCV= {Int. Conf. Comput. Vis.})

@String(ECCV= {Eur. Conf. Comput. Vis.})

@String(ICLR = {Int. Conf. Learn. Represent.})

@String(CVPR  = {CVPR})

@String(ICCV  = {ICCV})

@String(ECCV  = {ECCV})

@String(ICLR  = {ICLR})

@article{DAVIS,
  title={The 2017 davis challenge on video object segmentation},
  author={Pont-Tuset, Jordi and Perazzi, Federico and Caelles, Sergi and Arbel{\'a}ez, Pablo and Sorkine-Hornung, Alex and Van Gool, Luc},
  journal={arXiv preprint arXiv:1704.00675},
  year={2017}
}

@article{GAN,
  title={Generative adversarial nets},
  author={Goodfellow, Ian J and Pouget-Abadie, Jean and Mirza, Mehdi and Xu, Bing and Warde-Farley, David and Ozair, Sherjil and Courville, Aaron and Bengio, Yoshua},
  journal={Advances in Neural Information Processing Systems (NeurIPS)},
  volume={27},
  year={2014}
}

@inproceedings{LPIPS,
  title={The unreasonable effectiveness of deep features as a perceptual metric},
  author={Zhang, Richard and Isola, Phillip and Efros, Alexei A and Shechtman, Eli and Wang, Oliver},
  booktitle={Proceedings of the IEEE conference on computer vision and pattern recognition (CVPR)},
  pages={586--595},
  year={2018}
}

@inproceedings{StyleGAN2,
  title={Analyzing and improving the image quality of stylegan},
  author={Karras, Tero and Laine, Samuli and Aittala, Miika and Hellsten, Janne and Lehtinen, Jaakko and Aila, Timo},
  booktitle={Proceedings of the IEEE/CVF Conference on Computer Vision and Pattern Recognition (CVPR)},
  pages={8110--8119},
  year={2020}
}

@article{DDPM,
  title={Denoising diffusion probabilistic models},
  author={Ho, Jonathan and Jain, Ajay and Abbeel, Pieter},
  journal={Advances in Neural Information Processing Systems (NeurIPS)},
  volume={33},
  pages={6840--6851},
  year={2020}
}

@inproceedings{DDIM,
title={Denoising Diffusion Implicit Models},
author={Jiaming Song and Chenlin Meng and Stefano Ermon},
booktitle={International Conference on Learning Representations (ICLR)},
year={2021},
url={https://openreview.net/forum?id=St1giarCHLP}
}

@inproceedings{CLIP,
  title={Learning transferable visual models from natural language supervision},
  author={Radford, Alec and Kim, Jong Wook and Hallacy, Chris and Ramesh, Aditya and Goh, Gabriel and Agarwal, Sandhini and Sastry, Girish and Askell, Amanda and Mishkin, Pamela and Clark, Jack and others},
  booktitle={International Conference on Machine Learning (ICML)},
  pages={8748--8763},
  year={2021},
  organization={PmLR}
}

@inproceedings{LDM,
  title={High-resolution image synthesis with latent diffusion models},
  author={Rombach, Robin and Blattmann, Andreas and Lorenz, Dominik and Esser, Patrick and Ommer, Bj{\"o}rn},
  booktitle={Proceedings of the IEEE/CVF Conference on Computer Vision and Pattern Recognition (CVPR)},
  pages={10684--10695},
  year={2022}
}

@article{DALL-E-2,
  title={Hierarchical text-conditional image generation with clip latents},
  author={Ramesh, Aditya and Dhariwal, Prafulla and Nichol, Alex and Chu, Casey and Chen, Mark},
  journal={arXiv preprint arXiv:2204.06125},
  volume={1},
  number={2},
  pages={3},
  year={2022}
}

@article{Imagen,
  title={Photorealistic text-to-image diffusion models with deep language understanding},
  author={Saharia, Chitwan and Chan, William and Saxena, Saurabh and Li, Lala and Whang, Jay and Denton, Emily L and Ghasemipour, Kamyar and Gontijo Lopes, Raphael and Karagol Ayan, Burcu and Salimans, Tim and others},
  journal={Advances in Neural Information Processing Systems (NeurIPS)},
  volume={35},
  pages={36479--36494},
  year={2022}
}

@inproceedings{GLIDE,
  title={GLIDE: Towards photorealistic image generation and editing with text-guided diffusion models},
  author={Nichol, Alex and Dhariwal, Prafulla and Ramesh, Aditya and Shyam, Pranav and Mishkin, Pamela and McGrew, Bob and Sutskever, Ilya and Chen, Mark},
  booktitle={International Conference on Machine Learning (ICML)},
  year={2022}
}

@inproceedings{SDEdit,
      title={Sdedit: Guided image synthesis and editing with stochastic differential equations},
      author={Meng, Chenlin and He, Yutong and Song, Yang and Song, Jiaming and Wu, Jiajun and Zhu, Jun-Yan and Ermon, Stefano},
      booktitle={International Conference on Learning Representations (ICLR)},
      year={2022},
}

@inproceedings{LoRA,
  title={Lora: Low-rank adaptation of large language models.},
  author={Edward J Hu and Yelong Shen and Phillip Wallis and Zeyuan Allen-Zhu and Yuanzhi Li and Shean Wang and Lu Wang and Weizhu Chen},
  booktitle={International Conference on Learning Representations (ICLR)},
  year={2022}
}

@inproceedings{InstructPix2Pix,
  title={Instructpix2pix: Learning to follow image editing instructions},
  author={Brooks, Tim and Holynski, Aleksander and Efros, Alexei A},
  booktitle={Proceedings of the IEEE/CVF Conference on Computer Vision and Pattern Recognition (CVPR)},
  pages={18392--18402},
  year={2023}
}

@inproceedings{DreamBooth,
  title={Dreambooth: Fine tuning text-to-image diffusion models for subject-driven generation},
  author={Ruiz, Nataniel and Li, Yuanzhen and Jampani, Varun and Pritch, Yael and Rubinstein, Michael and Aberman, Kfir},
  booktitle={Proceedings of the IEEE/CVF Conference on Computer Vision and Pattern Recognition (CVPR)},
  pages={22500--22510},
  year={2023}
}

@inproceedings{DragGAN,
  title={Drag your gan: Interactive point-based manipulation on the generative image manifold},
  author={Pan, Xingang and Tewari, Ayush and Leimk{\"u}hler, Thomas and Liu, Lingjie and Meka, Abhimitra and Theobalt, Christian},
  booktitle={ACM SIGGRAPH 2023 Conference Proceedings},
  pages={1--11},
  year={2023}
}

@article{DIFT,
  title={Emergent correspondence from image diffusion},
  author={Tang, Luming and Jia, Menglin and Wang, Qianqian and Phoo, Cheng Perng and Hariharan, Bharath},
  journal={Advances in Neural Information Processing Systems (NeurIPS)},
  volume={36},
  pages={1363--1389},
  year={2023}
}

@article{DiffusionHyperfeatures,
  title={Diffusion hyperfeatures: Searching through time and space for semantic correspondence},
  author={Luo, Grace and Dunlap, Lisa and Park, Dong Huk and Holynski, Aleksander and Darrell, Trevor},
  journal={Advances in Neural Information Processing Systems (NeurIPS)},
  volume={36},
  pages={47500--47510},
  year={2023}
}

@inproceedings{ControlNet,
  title={Adding conditional control to text-to-image diffusion models},
  author={Zhang, Lvmin and Rao, Anyi and Agrawala, Maneesh},
  booktitle={Proceedings of the IEEE/CVF International Conference on Computer Vision (ICCV)},
  pages={3836--3847},
  year={2023}
}

@inproceedings{DragDiffusion,
  title={Dragdiffusion: Harnessing diffusion models for interactive point-based image editing},
  author={Shi, Yujun and Xue, Chuhui and Liew, Jun Hao and Pan, Jiachun and Yan, Hanshu and Zhang, Wenqing and Tan, Vincent YF and Bai, Song},
  booktitle={Proceedings of the IEEE/CVF Conference on Computer Vision and Pattern Recognition (CVPR)},
  pages={8839--8849},
  year={2024}
}

@inproceedings{DragonDiffusion,
  title={Dragondiffusion: Enabling drag-style manipulation on diffusion models},
  author={Mou, Chong and Wang, Xintao and Song, Jiechong and Shan, Ying and Zhang, Jian},
  booktitle={International Conference on Learning Representations (ICLR)},
  year={2024}
}

@inproceedings{FreeDrag,
  title={Freedrag: Feature dragging for reliable point-based image editing},
  author={Ling, Pengyang and Chen, Lin and Zhang, Pan and Chen, Huaian and Jin, Yi and Zheng, Jinjin},
  booktitle={Proceedings of the IEEE/CVF Conference on Computer Vision and Pattern Recognition (CVPR)},
  pages={6860--6870},
  year={2024}
}

@inproceedings{SDEDrag,
   title={The Blessing of Randomness: SDE Beats ODE in General Diffusion-based Image Editing},
   author={Nie, Shen and Guo, Hanzhong Allan and Lu, Cheng and Zhou, Yuhao and Zheng, Chenyu and Li, Chongxuan},
   booktitle={International Conference on Learning Representations (ICLR)},
   year={2024}
}

@inproceedings{DragNoise,
  title={Drag your noise: Interactive point-based editing via diffusion semantic propagation},
  author={Liu, Haofeng and Xu, Chenshu and Yang, Yifei and Zeng, Lihua and He, Shengfeng},
  booktitle={Proceedings of the IEEE/CVF Conference on Computer Vision and Pattern Recognition (CVPR)},
  pages={6743--6752},
  year={2024}
}

@inproceedings{DiffEditor,
  title={Diffeditor: Boosting accuracy and flexibility on diffusion-based image editing},
  author={Mou, Chong and Wang, Xintao and Song, Jiechong and Shan, Ying and Zhang, Jian},
  booktitle={Proceedings of the IEEE/CVF Conference on Computer Vision and Pattern Recognition (CVPR)},
  pages={8488--8497},
  year={2024}
}

@inproceedings{EasyDrag,
  title={Easydrag: Efficient point-based manipulation on diffusion models},
  author={Hou, Xingzhong and Liu, Boxiao and Zhang, Yi and Liu, Jihao and Liu, Yu and You, Haihang},
  booktitle={Proceedings of the IEEE/CVF Conference on Computer Vision and Pattern Recognition (CVPR)},
  pages={8404--8413},
  year={2024}
}

@article{FastDrag,
  title={Fastdrag: Manipulate anything in one step},
  author={Zhao, Xuanjia and Guan, Jian and Fan, Congyi and Xu, Dongli and Lin, Youtian and Pan, Haiwei and Feng, Pengming},
  journal={Advances in Neural Information Processing Systems (NeurIPS)},
  volume={37},
  pages={74439--74460},
  year={2024}
}

@inproceedings{InstantDrag,
  title={Instantdrag: Improving interactivity in drag-based image editing},
  author={Shin, Joonghyuk and Choi, Daehyeon and Park, Jaesik},
  booktitle={SIGGRAPH Asia 2024 Conference Papers},
  pages={1--10},
  year={2024}
}

@article{AdaptiveDrag,
  title={AdaptiveDrag: Mask-Free Point-Based Image Editing with Editable Region Localization},
  author={Chen, Yining and Wang, Qi and Zhu, Hao and Lin, Hongxu and Xu, Yibing},
  journal={arXiv preprint arXiv:2410.12696},
  year={2024}
}

@inproceedings{StableDrag,
  title={Stabledrag: Stable dragging for point-based image editing},
  author={Cui, Yutao and Zhao, Xiaotong and Zhang, Guozhen and Cao, Shengming and Ma, Kai and Wang, Limin},
  booktitle={European Conference on Computer Vision (ECCV)},
  pages={340--356},
  year={2024},
  organization={Springer}
}

@inproceedings{ReadoutGuidance,
  title={Readout guidance: Learning control from diffusion features},
  author={Luo, Grace and Darrell, Trevor and Wang, Oliver and Goldman, Dan B and Holynski, Aleksander},
  booktitle={Proceedings of the IEEE/CVF Conference on Computer Vision and Pattern Recognition (CVPR)},
  pages={8217--8227},
  year={2024}
}

@inproceedings{GoodDrag,
    title={GoodDrag: Towards Good Practices for Drag Editing with Diffusion Models},
    author={Zewei Zhang and Huan Liu and Jun Chen and Xiangyu Xu},
    booktitle={International Conference on Learning Representations (ICLR)},
    year={2025},
}

@inproceedings{LightningDrag,
     title={LightningDrag: Lightning Fast and Accurate Drag-based Image Editing Emerging from Videos},
     author = {Shi, Yujun and Liew, Jun Hao and Yan, Hanshu and Tan, Vincent YF and Feng, Jiashi},
     booktitle={International Conference on Machine Learning (ICML)},
     year={2025}
}

@inproceedings{DragText,
  title={Dragtext: Rethinking text embedding in point-based image editing},
  author={Choi, Gayoon and Jeong, Taejin and Hong, Sujung and Hwang, Seong Jae},
  booktitle={2025 IEEE/CVF Winter Conference on Applications of Computer Vision (WACV)},
  pages={441--450},
  year={2025},
  organization={IEEE}
}

@inproceedings{SAM2,
  title={Sam 2: Segment anything in images and videos},
  author={Ravi, Nikhila and Gabeur, Valentin and Hu, Yuan-Ting and Hu, Ronghang and Ryali, Chaitanya and Ma, Tengyu and Khedr, Haitham and R{\"a}dle, Roman and Rolland, Chloe and Gustafson, Laura and Mintun, Eric and Pan, Junting and Alwala, Kalyan Vasudev and Carion, Nicolas and Wu, Chao-Yuan and Girshick, Ross and Doll{\'a}r, Piotr and Feichtenhofer, Christoph},
  booktitle={International Conference on Machine Learning (ICML)},
  year={2025}
}

@inproceedings{GDrag,
  title={Gdrag: Towards general-purpose interactive editing with anti-ambiguity point diffusion},
  author={Lin, Xiaojian and Li, Hanhui and Cheng, Yuhao and Yan, Yiqiang and Liang, Xiaodan},
  booktitle={The Thirteenth International Conference on Learning Representations (ICLR)},
  year={2025}
}

@inproceedings{DADG,
  title={Training-free Dense-Aligned Diffusion Guidance for Modular Conditional Image Synthesis},
  author={Wang, Zixuan and Peng, Duo and Chen, Feng and Yang, Yuwei and Lei, Yinjie},
  booktitle={Proceedings of the IEEE/CVF Conference on Computer Vision and Pattern Recognition (CVPR)},
  pages={13135--13145},
  year={2025}
}

@article{EEdit,
  title={Eedit: Rethinking the spatial and temporal redundancy for efficient image editing},
  author={Yan, Zexuan and Ma, Yue and Zou, Chang and Chen, Wenteng and Chen, Qifeng and Zhang, Linfeng},
  journal={arXiv preprint arXiv:2503.10270},
  year={2025}
}

@inproceedings{FlowDrag,
  title={Flowdrag: 3d-aware drag-based image editing with mesh-guided deformation vector flow fields},
  author={Koo, Gwanhyeong and Yoon, Sunjae and Lee, Younghwan and Hong, Ji Woo and Yoo, Chang D},
  booktitle={International Conference on Machine Learning (ICML)},
  year={2025}
}

@inproceedings{DragLoRA,
  title={DragLoRA: Online Optimization of LoRA Adapters for Drag-based Image Editing in Diffusion Model},
  author={Xia, Siwei and Sun, Li and Sun, Tiantian and Li, Qingli},
  booktitle={International Conference on Machine Learning (ICML)},
  year={2025}
}
}
\clearpage


\twocolumn[{%
\renewcommand\twocolumn[1][]{#1}
\begin{center}
    \centering
    \maketitlesupplementary
    \captionsetup{type=figure}

\end{center}
}]

\section{Supplementary Material}\label{sec:supplementary_material}

\subsection{Improper Mask and Prompt}

\begin{figure*}[htbp]
    \centering
    \includegraphics[width=0.85\linewidth]{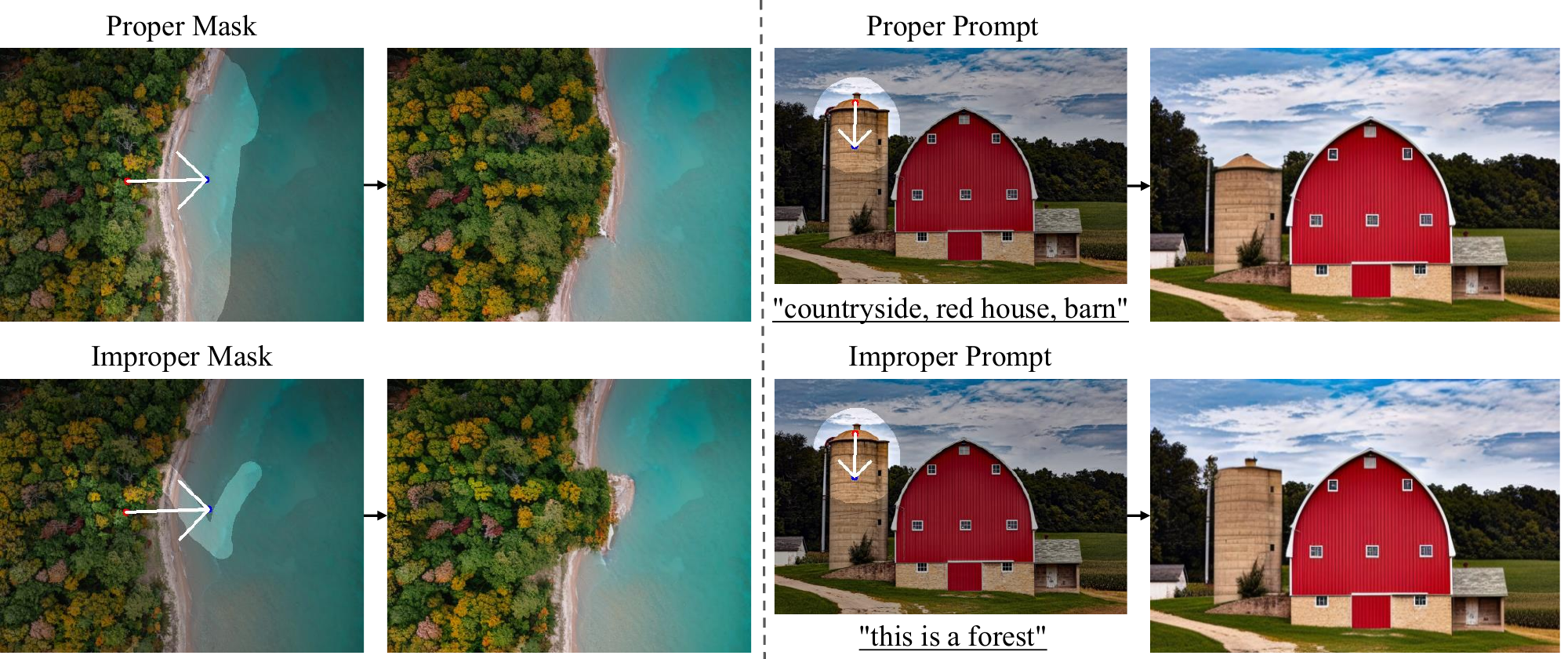}
    \caption{\textbf{Improper Mask and Prompt}
    }
    \label{fig:improper}
\end{figure*}

Figure~\ref{fig:improper} illustrates how poorly drawn mask or irrelevant prompt can cause distortion or semantic failure. Designing such inputs is often nontrivial and error-prone. This motivates our manual mask-free and prompt-free approach.

\subsection{Unexpected Benefits Without Mask and Prompt}

\begin{figure*}[htbp]
    \centering
    \includegraphics[width=0.6\linewidth]{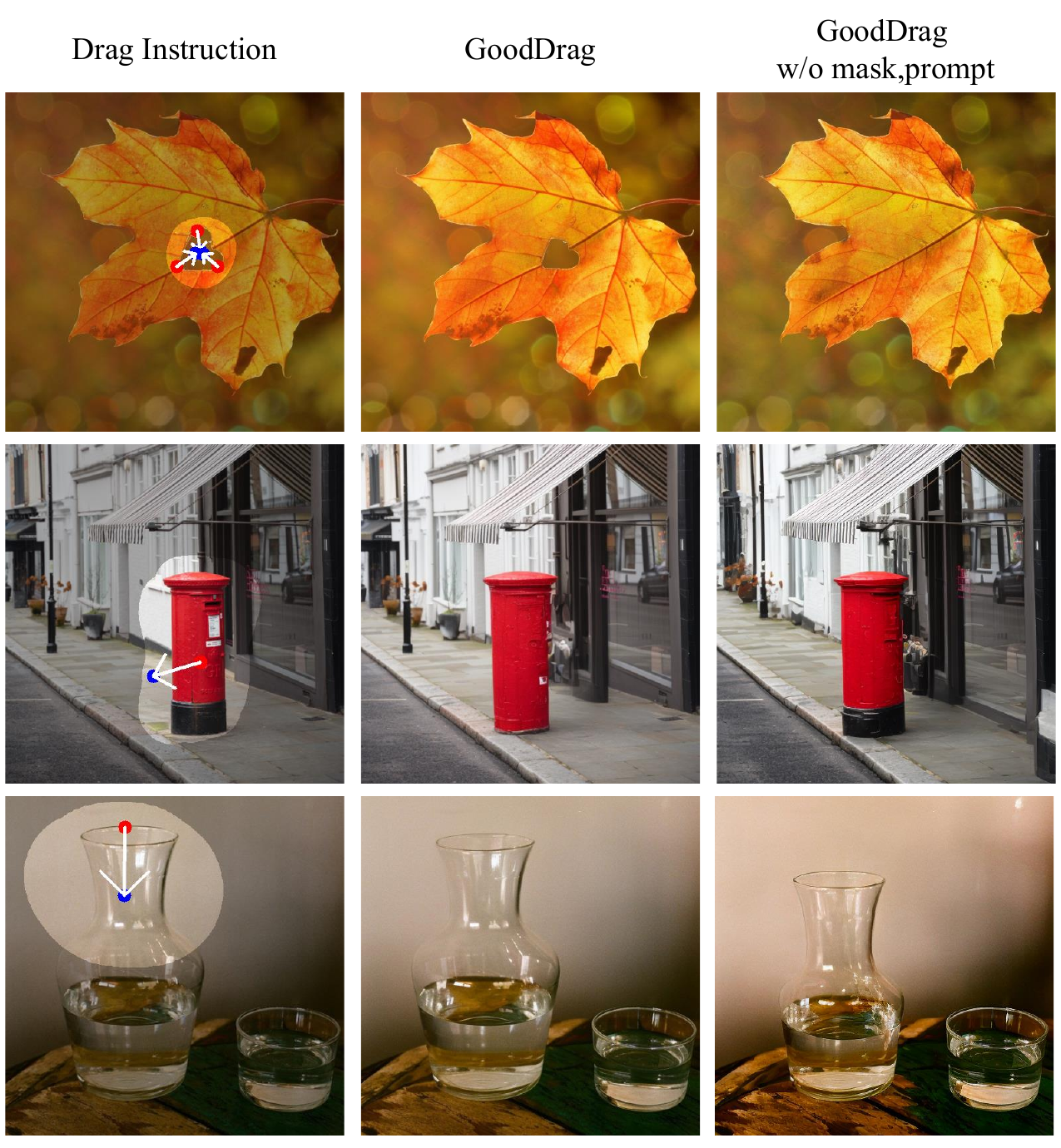}
    \caption{\textbf{Unexpected Benefits Without Mask and Prompt}
    }
    \label{fig:benefit_no_mask_prompt}
\end{figure*}

In rare cases, removing the mask or prompt surprisingly improves results. Figure~\ref{fig:benefit_no_mask_prompt} shows edits that are better or more natural without mask and prompt. This experiment is conducted using the GoodDrag~\cite{GoodDrag} method.

\subsection{Readout Network Architecture}

\begin{figure*}[htbp]
    \centering
    \includegraphics[width=0.95\linewidth]{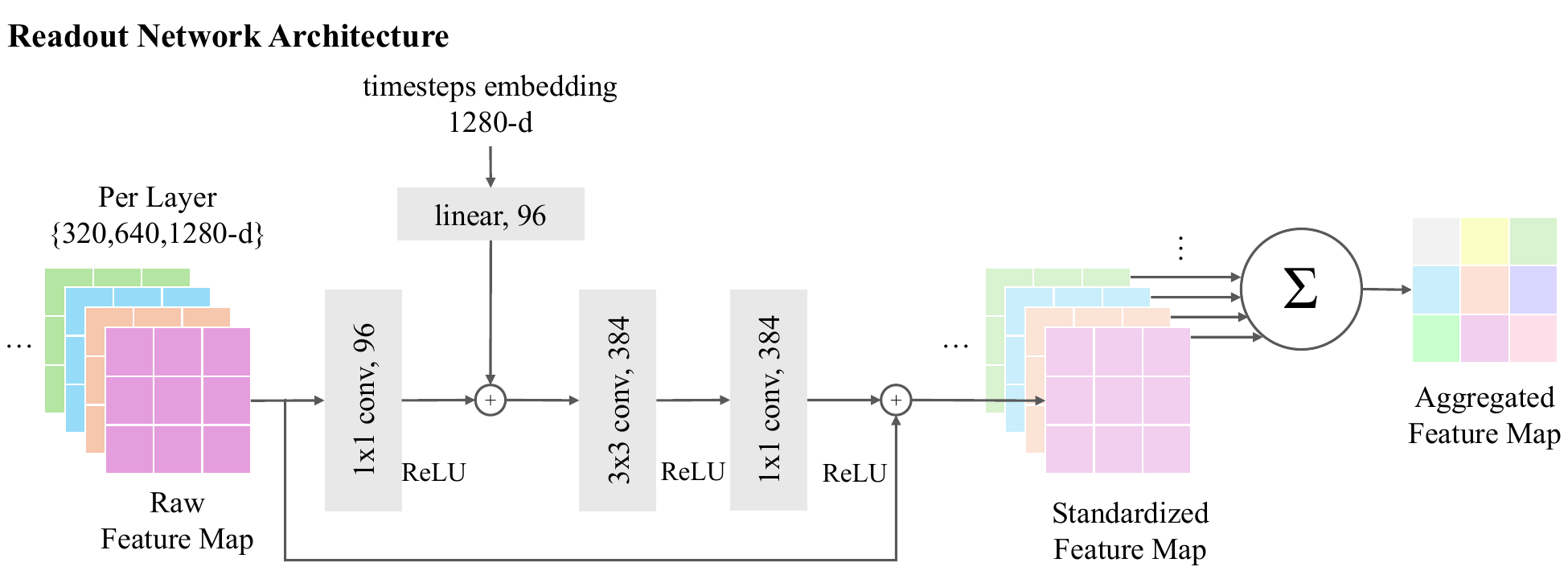}
    \caption{\textbf{Readout Network Architecture}
    }
    \label{fig:readout_arch}
\end{figure*}

Our readout module directly follows the architecture introduced in Readout Guidance~\cite{DreamBooth} (see Figure~\ref{fig:readout_arch}). We use the same aggregation network to extract intermediate features from the decoder, where each decoder feature is first passed through a bottleneck layer to standardize the channel size. These bottleneck layers are made timestep-conditional by adding projected timestep embeddings, obtained from the pretrained U-Net’s timestep encoding. The standardized features are then aggregated via a learned weighted sum. We adopt this architecture without structural changes, using it as a guidance signal for feature alignment in our drag-editing task.

\subsection{Comparison with InstantDrag}

\begin{figure*}[htbp]
    \centering
    \includegraphics[width=0.95\linewidth]{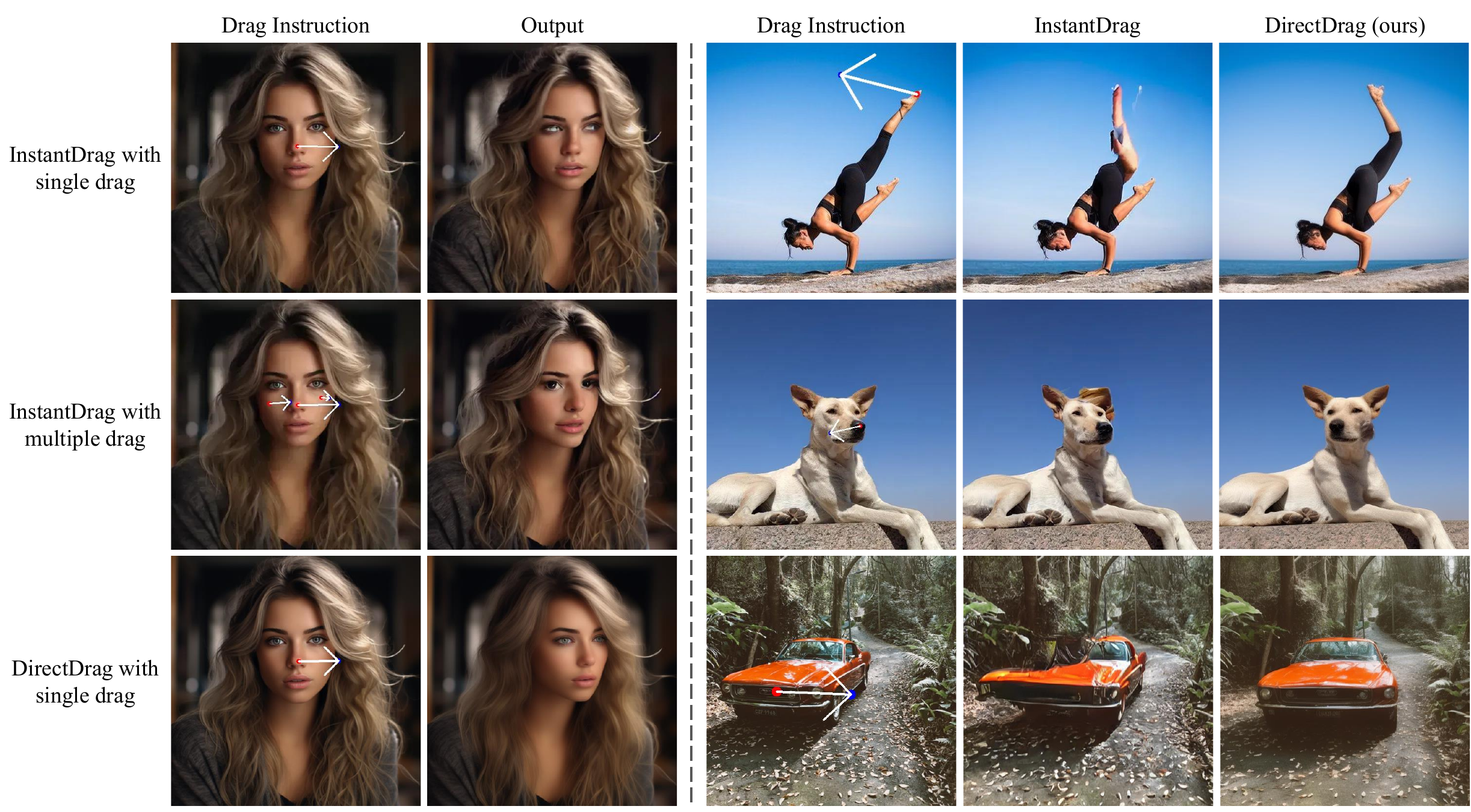}
    \caption{\textbf{Comparison with InstantDrag~\cite{InstantDrag}} Left: InstantDrag requires multiple drag instructions to rotate the face, while \ours achieves similar results with a single drag instruction. Right: InstantDrag often produces unstable or distorted results, while our method yields more faithful and coherent outputs.
    }
    \label{fig:instantdrag_compare}
\end{figure*}

See Figure~\ref{fig:instantdrag_compare} for visual comparisons with InstantDrag~\cite{InstantDrag}.

\subsection{Extended Qualitative Comparison}

\begin{figure*}[htbp]
    \centering
    \includegraphics[width=0.78\linewidth]{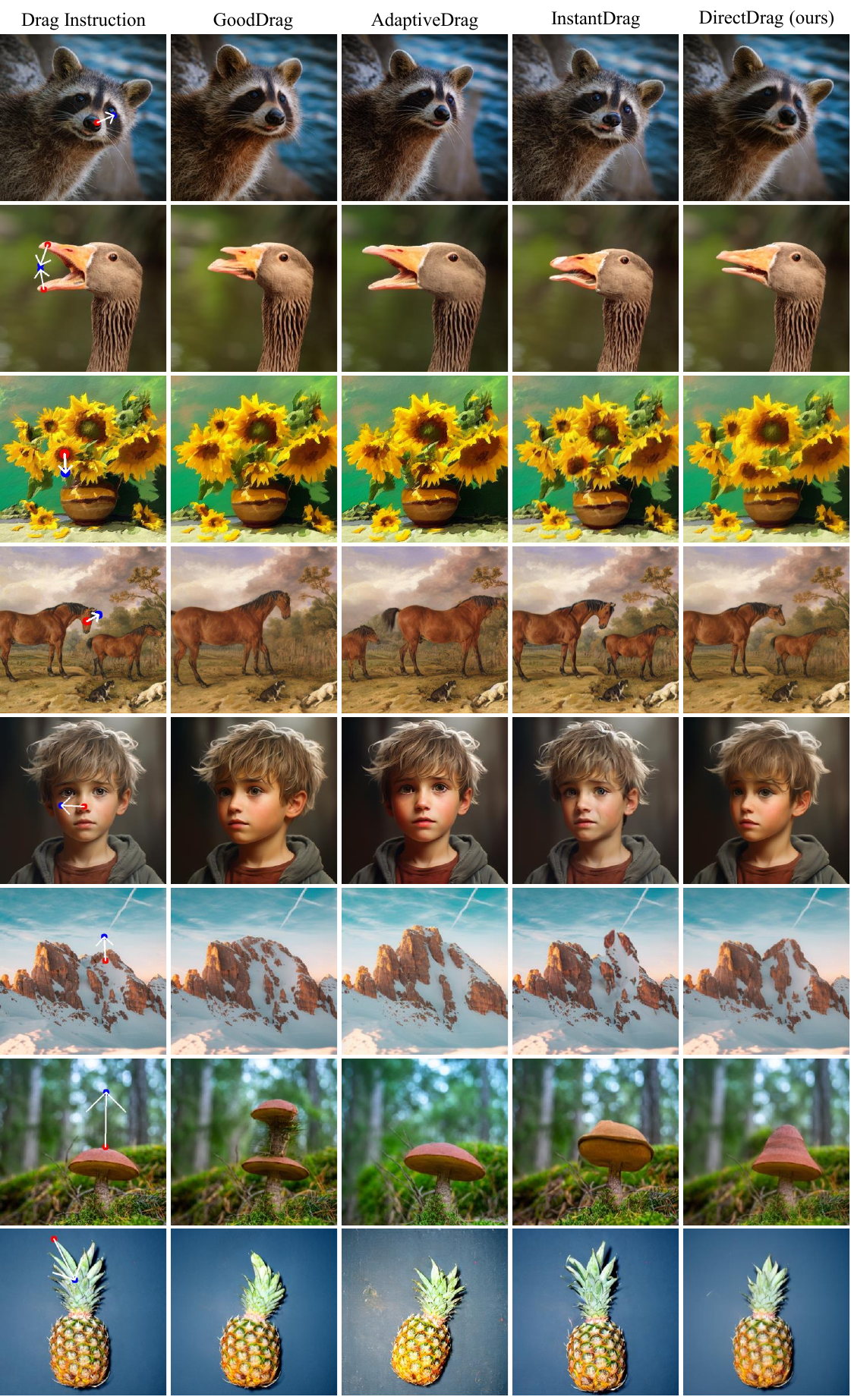}
    \caption{\textbf{Extended Qualitative Comparison}
    }
    \label{fig:additional_compare1}
\end{figure*}

\begin{figure*}[htbp]
    \centering
    \includegraphics[width=0.78\linewidth]{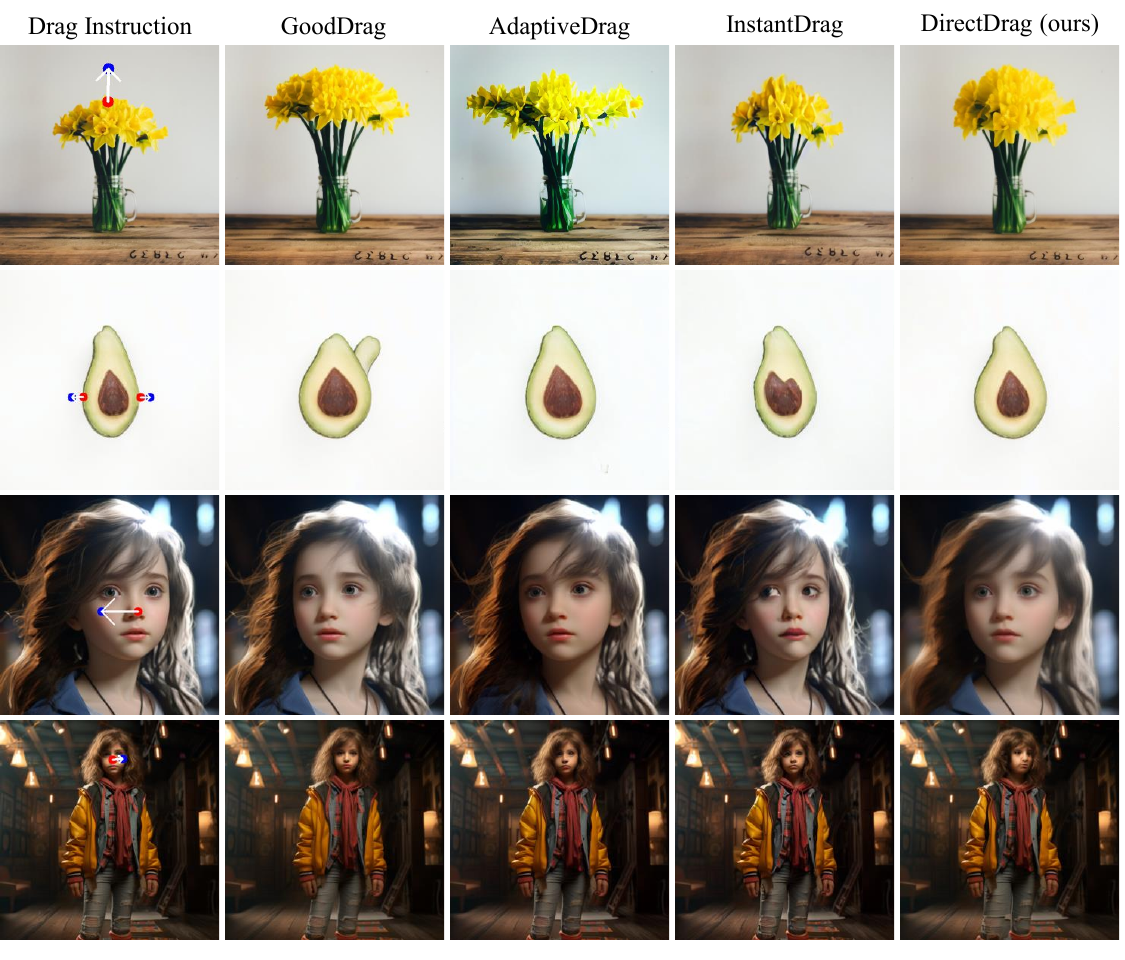}
    \caption{\textbf{Extended Qualitative Comparison}
    }
    \label{fig:additional_compare2}
\end{figure*}

We present additional comparisons with prior drag-based methods to highlight differences in fidelity and accuracy. See Figure~\ref{fig:additional_compare1} and Figure~\ref{fig:additional_compare2}.

\subsection{Extended Qualitative Examples}

\begin{figure*}[htbp]
    \centering
    \includegraphics[width=0.95\linewidth]{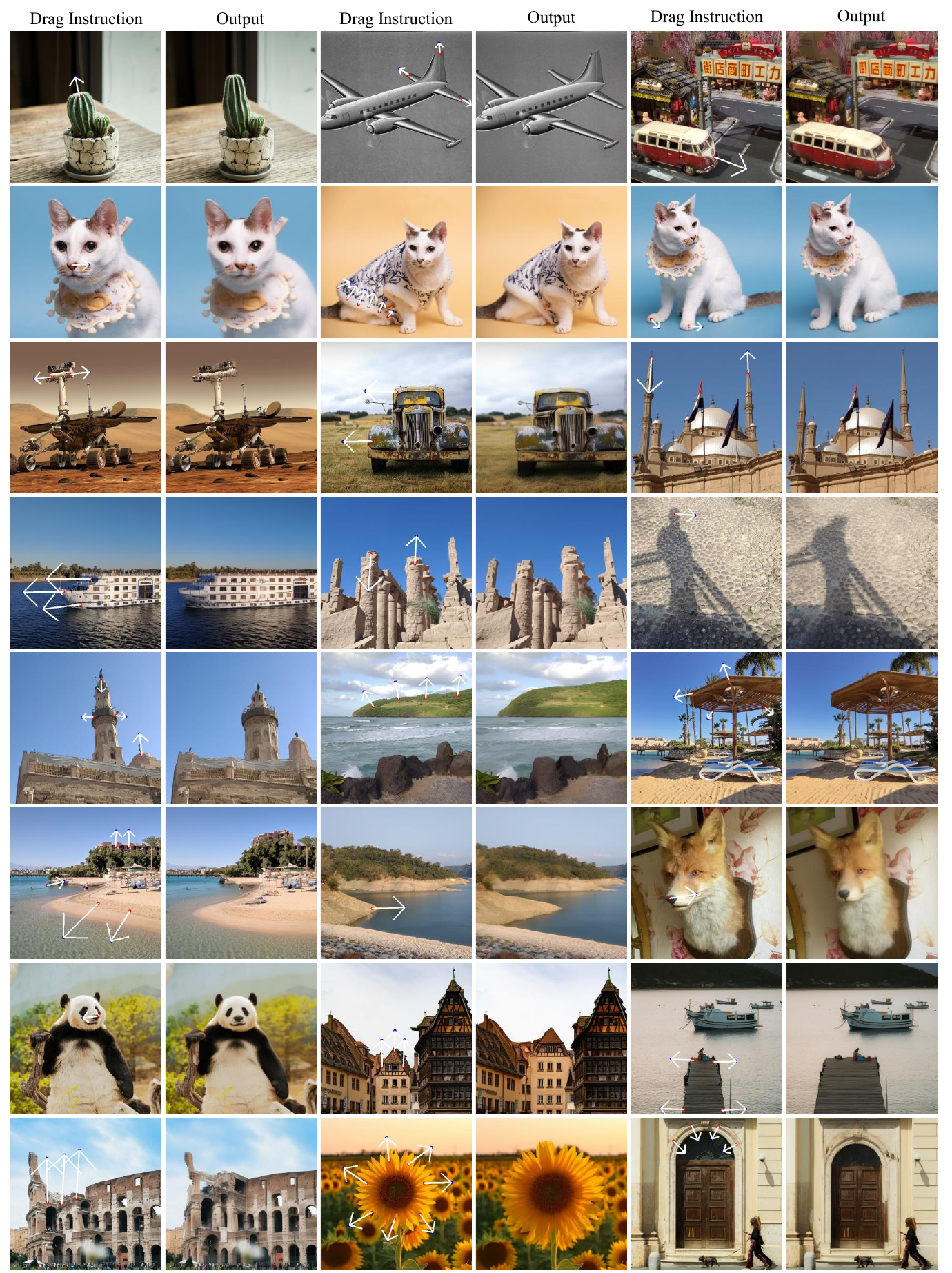}
    \caption{\textbf{Extended Qualitative Examples}
    }
    \label{fig:additional_result}
\end{figure*}

We showcase more qualitative results produced by \ours~across diverse scenes and manipulation tasks. See Figure~\ref{fig:additional_result}.

\subsection{Qualitative Results of the Ablation Study}

\begin{figure*}[htbp]
    \centering
    \includegraphics[width=0.78\linewidth]{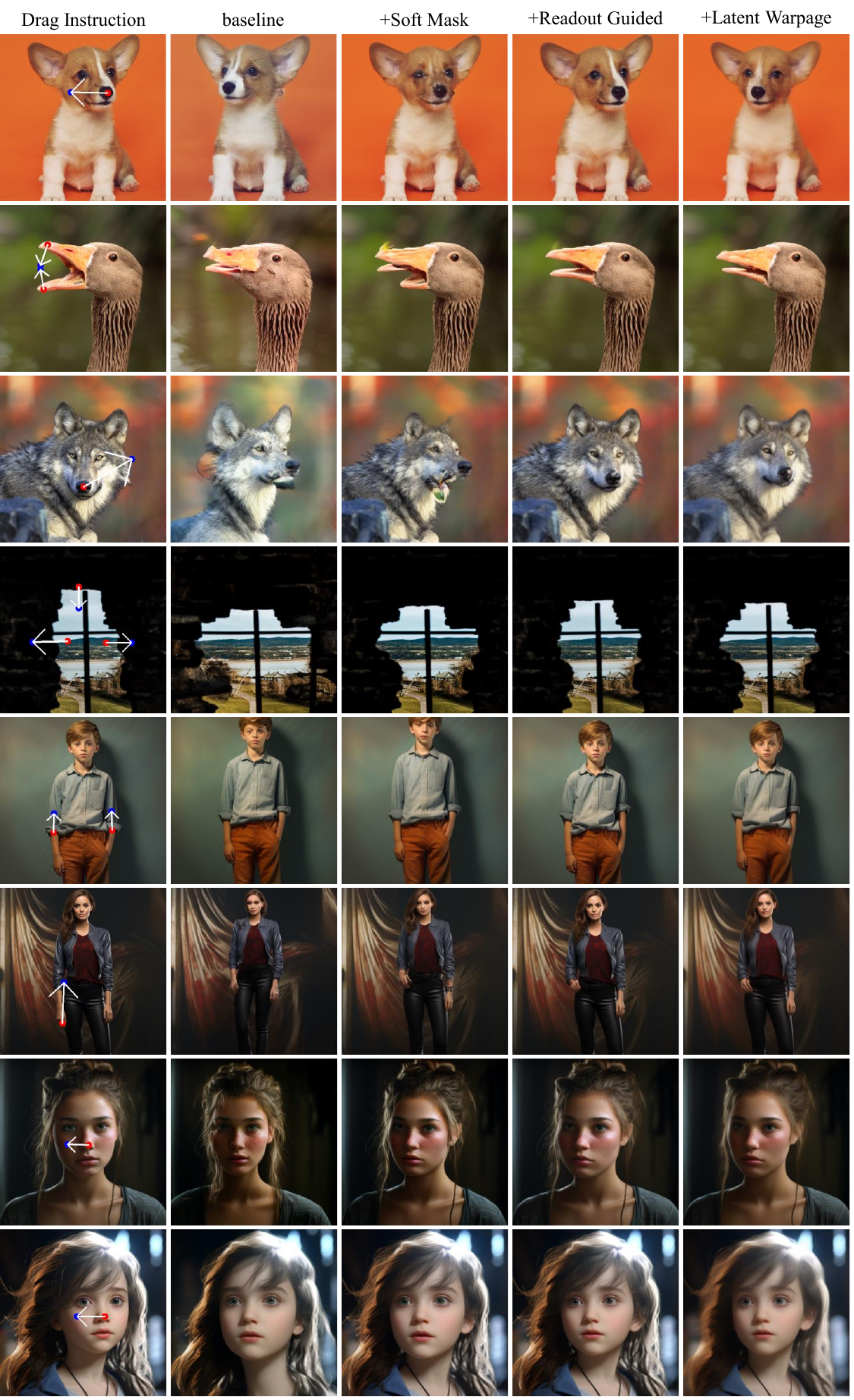}
    \caption{\textbf{Qualitative Results of the Ablation Study}
    }
    \label{fig:ablation_example1}
\end{figure*}

\begin{figure*}[htbp]
    \centering
    \includegraphics[width=0.78\linewidth]{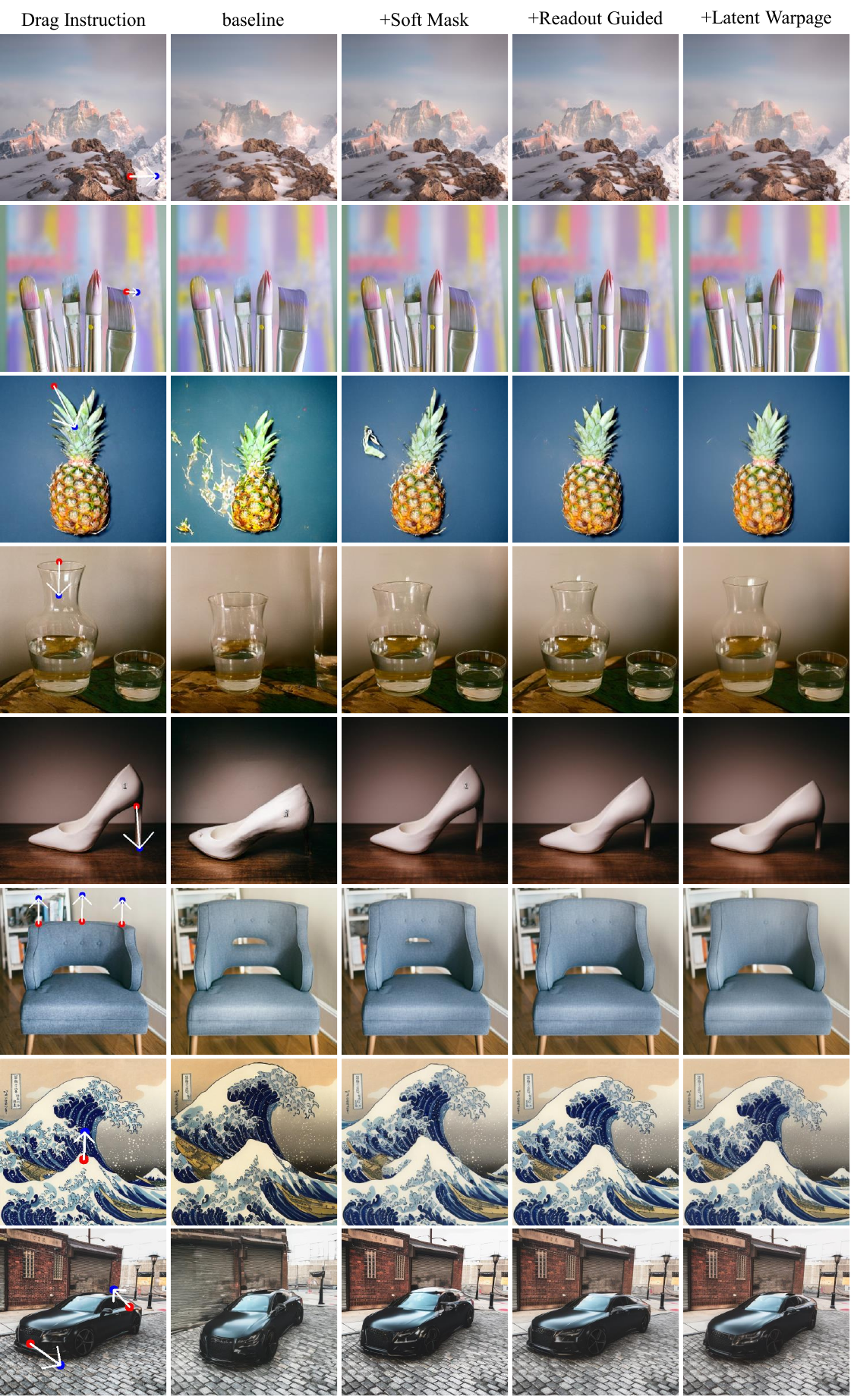}
    \caption{\textbf{Qualitative Results of the Ablation Study}
    }
    \label{fig:ablation_example2}
\end{figure*}

To better understand the role of each component, we visualize editing results under different ablation settings. See Figure~\ref{fig:ablation_example1} and Figure~\ref{fig:ablation_example2}.

\subsection{Limitations and Example}

\begin{figure*}[htbp]
    \centering
    \includegraphics[width=0.78\linewidth]{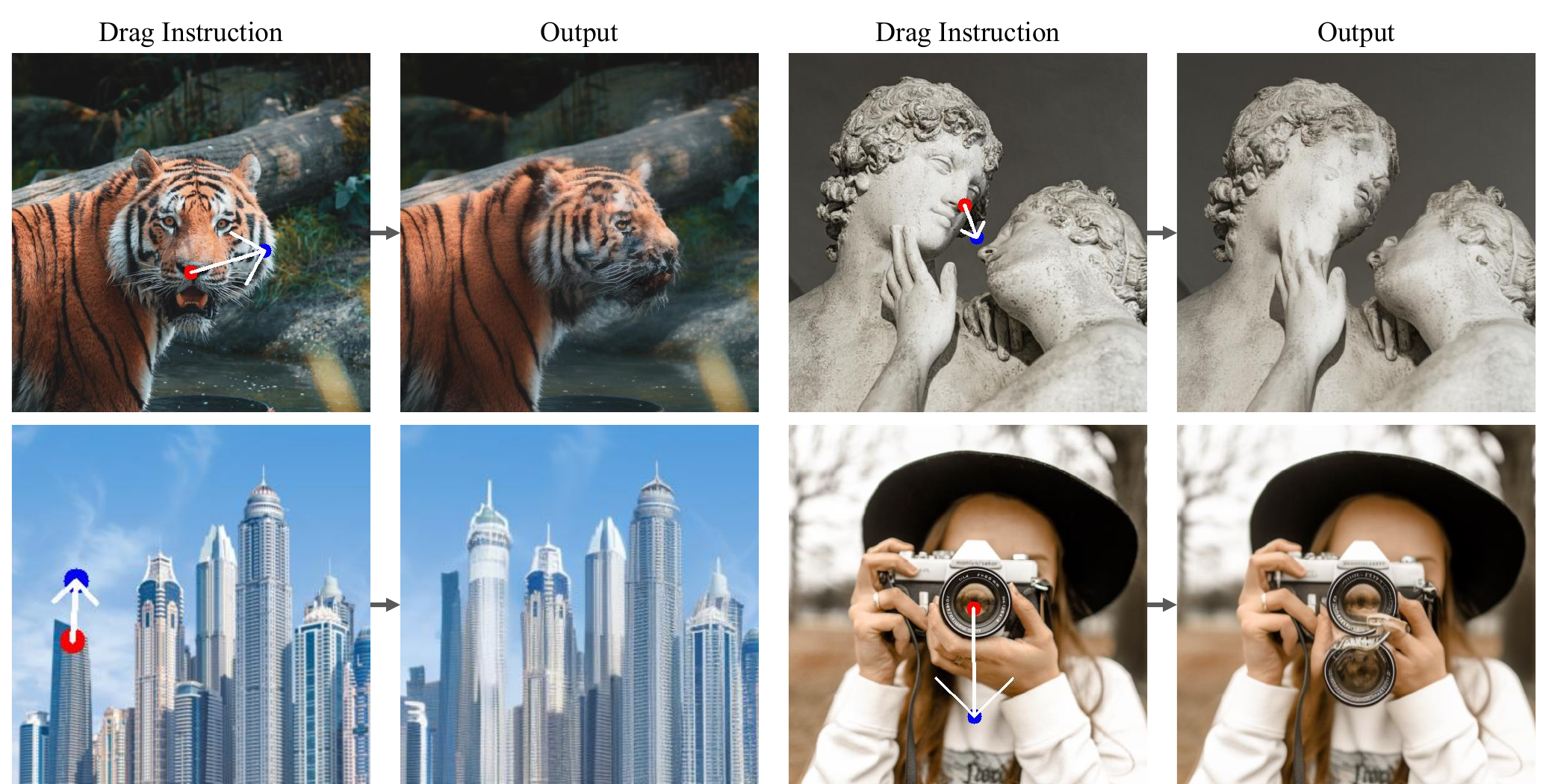}
    \caption{\textbf{Qualitative Results of Limitations}
    }
    \label{fig:limitation_example}
\end{figure*}

In some cases, our method may over-preserve visual fidelity, resulting in insufficient deformation. Additionally, strong geometric warping can occasionally cause texture detail loss. Furthermore, our method is still less aligned with human intent compared to manual dragging. See Figure~\ref{fig:limitation_example}.

\clearpage

\clearpage

\clearpage

\clearpage

\clearpage

\clearpage

\clearpage

\clearpage

\end{document}